%% file: main.tex
\documentclass[11pt,a4]{article}
\usepackage[top=1in,left=1in,footskip=1in,marginparwidth=1in]{geometry}

\usepackage{fancyhdr}
\usepackage{hyperref}
\hypersetup{
    unicode=false,          % non-Latin characters in Acrobat’s bookmarks
    pdftoolbar=true,        % show Acrobat’s toolbar?
    pdfmenubar=true,        % show Acrobat’s menu?
    pdffitwindow=false,     % window fit to page when opened
    pdfstartview={FitH},    % fits the width of the page to the window
    pdftitle={},    % title
    pdfauthor={},     % author
    pdfsubject={},   % subject of the document
    pdfcreator={},   % creator of the document
    pdfproducer={}, % producer of the document
    pdfkeywords={key1} {key2} {key3}, % list of keywords
    pdfnewwindow=true,      % links in new window
    colorlinks=true,       % false: boxed links; true: colored links
    linkcolor=blue,          % color of internal links (change box color with linkbordercolor)
    citecolor=blue,        % color of links to bibliography
    filecolor=blue,      % color of file links
    urlcolor=blue           % color of external links
}
\usepackage[utf8]{inputenc}
\usepackage{tcolorbox}

\usepackage{cite}

\usepackage{nameref,hyperref}

\usepackage[right]{lineno}

\usepackage{microtype}
\DisableLigatures[f]{encoding = *, family = * }

\usepackage{parskip}

\usepackage{changepage}

\usepackage[aboveskip=1pt,labelfont=bf,labelsep=period,singlelinecheck=off]{caption}

\makeatletter
\renewcommand{\@biblabel}[1]{\quad#1.}
\makeatother

\usepackage{lastpage,fancyhdr,graphicx}
\usepackage{epstopdf}
\fancyheadoffset[L]{2.25in}
\fancyfootoffset[L]{2.25in}

\usepackage{color}

\definecolor{Gray}{gray}{.25}

\usepackage{graphicx}

\usepackage{sidecap}

\usepackage{textcomp}
\usepackage{blindtext}
\usepackage{tcolorbox}
\usepackage{graphicx}
\usepackage{enumitem}
\usepackage{multirow}
\usepackage{subfigure}
\begin{document}
\vspace*{0.35in}

% title goes here:
\begin{flushleft}
{\Large
\textbf\newline{\Huge Artificial intelligence to advance Earth observation: a perspective}
}
\newline
% authors go here:
\\
Devis Tuia\textsuperscript{1,*}, 
Konrad Schindler\textsuperscript{2},  
Beg{\"u}m Demir\textsuperscript{3,4},  
Xiao Xiang Zhu\textsuperscript{5},  
Mrinalini Kochupillai\textsuperscript{5}, 
Sa\v{s}o D\v{z}eroski\textsuperscript{6}, 
Jan N. van Rijn\textsuperscript{7}, 
Holger H. Hoos\textsuperscript{7,13},
Fabio Del Frate\textsuperscript{8}, 
Mihai Datcu\textsuperscript{9,10}, 
Volker Markl\textsuperscript{3,4}, 
Bertrand Le Saux\textsuperscript{11}, 
Rochelle Schneider\textsuperscript{11}, 
Gustau Camps-Valls\textsuperscript{12}
\\
\bigskip
\bf{$^1$} {Ecole Polytechnique Fédérale de Lausanne (EPFL), Switzerland}\\
\bf{$^2$} {ETH Zurich, Switzerland}\\
\bf{$^3$} {Technische Universit{\"a}t Berlin, Germany}\\
\bf{$^{4}$} {BIFOLD - Berlin Institute for the Foundations of Learning and Data, Germany}\\
\bf{$^5$} {Technische Universit{\"a}t München, Germany}\\
\bf{$^6$} {Jozef Stefan Institute, Ljubljana, Slovenia}\\
\bf{$^7$} {Leiden University, the Netherlands}\\
\bf{$^{13}$} {RWTH Aachen University, Germany}\\
\bf{$^8$} {University of Rome ``Tor Vergata'', Italy}\\
\bf{$^9$} {German Aerospace Center (DLR), Germany}\\
\bf{$^{10}$} {University Politehnica of Bucharest (UPB), Romania}\\
\bf{$^{11}$} {European Space Agency (ESA) $\Phi$-lab, Italy}\\
\bf{$^{12}$} {Universitat de Val{\`e}ncia, Spain}\\

\bigskip
* Corresponding author: devis.tuia@epfl.ch. 

\vspace{1cm}

This article is a preprint of a paper accepted at IEEE Geoscience and Remote Sensing Magazine (DOI: 10.1109/MGRS.2024.3425961). Official published version available at: \url{https://ieeexplore.ieee.org/document/10669817}.

\end{flushleft}

\clearpage
\tableofcontents

\clearpage
\addcontentsline{toc}{section}{Introduction}

\begin{abstract}
Earth observation is increasingly used for mapping and monitoring processes occurring at the surface of Earth. Data acquired by satellites nowadays allow us to have a global view, consistent in time, of the state of our forests, oceans, and growing urban areas. However, such a wealth of data has little value without appropriate processing chains able to convert the pixel values to information useful for decision makers. \\
Recently, machine learning has seen fast advances -- especially thanks to the rise of deep learning methodologies -- and is increasingly deployed in Earth observation image processing systems. The ever-growing models from computer vision and natural language processing have inspired developments in remote sensing, and new approaches are constantly proposed in the field. However, despite their impressive results, the ever-growing mass of approaches and solutions makes it complicated to have a holistic overview and to know the most promising approaches from the field. \\
In this paper, we aim to fill this knowledge gap and propose to review the thriving ecosystem focusing on developing AI models for Earth observation,its recent trends, and sketch potential pathways for future advances. 
\end{abstract}

\section*{{Introduction}}\label{sec:intro}

EEarth observation (EO) is a prime instrument for monitoring land and ocean processes, studying the dynamics at work, and taking the pulse of our planet.  A large variety of sensor data (active / passive / of many resolutions) are nowadays accessible to researchers, agencies, and the general public. However, a final barrier remains the need for technology to convert the enormous quantities of raw EO data generated into the valuable information necessary for making decisions and taking concrete action. The ability to extract meaningful information from raw EO data is an essential prerequisite to achieving significant impact, be it in terms of monitoring or documenting (e.g. progress towards the United Nations' sustainable development goals), predicting and issuing timely warnings (e.g. related to future natural disasters~\cite{RNY009} and need for emergency evacuations), or projecting the effects of human actions and natural processes on nature and society~\cite{RNY006,RNY011,RNY012}.

From the beginning of EO research, computer science and signal processing techniques have been important for extracting meaningful information from raw data. We have observed rapid technological advances in these fields over the last few years, mainly owing to the increasing use of artificial intelligence (AI) methods -- and in particular, techniques from machine learning (ML) in the context of Earth observation. Notably, 
the use of deep learning (DL) and other ML techniques has had a broad and significant impact on EO and remote sensing \cite{zhu2017deep,persello22deep}, 
where they are used across entire processing chains, from data compression and transmission to image recognition and predictions of environmental variables (land cover, land use, biomass, etc.). 
Similarly, ML techniques are 
increasingly widely used in many areas of environmental science~\cite{Cam20b}{, with potential key roles to play to support decision making in areas such as food security, environmental health or biodiversity monitoring}.

However, while classical processing tasks, such as land cover classification, seem to have reached a certain degree of maturity, partially due to the better integration of EO and ML methods, substantial work remains to render these approaches truly useful for users and society. In a recent article, an agenda for future technical achievements required in this context has been proposed~\cite{Tui20grsm}, which triggered this effort from AI4EO scholars and the European Space Agency $\Phi$-Lab.  
Accordingly, with this overview article, we aim to explore and discuss in more detail the development and use of AI methods in the field of EO, with an emphasis on ML techniques, and to inspire the EO community, thus to realise the transformative advances made possible by these techniques.

\begin{figure*}[ht!]
    \centering
    \includegraphics[width=0.85\textwidth]{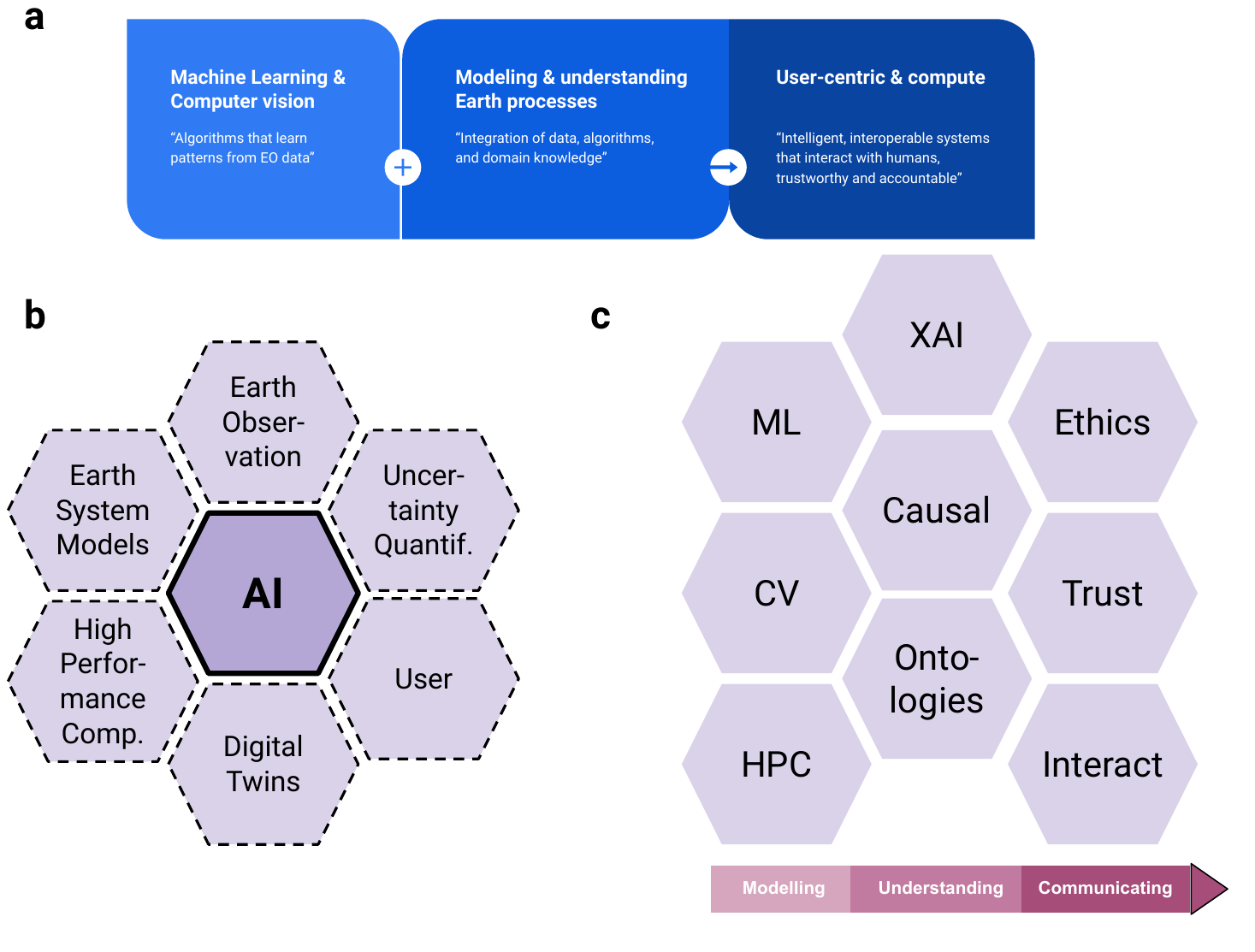}
    \caption{Conceptual overview of this perspective paper: (a) different levels of algorithms emerge from the areas of machine learning (ML) and interact with computer vision (CV), computer science, and statistics to learn patterns and associations from observational data. The models must integrate domain knowledge and biogeophysical constraints to advance in the modeling and understanding the Earth's processes. The ultimate goal is to provide intelligent, interoperable, actionable, trustworthy, robust systems whose decisions should be accountable. 
    (b) The field of AI-- and specifically, the area of ML within it -- interacts (and is embedded into) several systems to realise such ambitious goals, from high-performance computing platforms in digital twins to Earth system model simulations and products, a wide range of Earth observation data, the characterisation and quantification of uncertainty, and the (active) role of the users. 
    (c) The processing chain in AI goes from modeling (e.g. classification, detection, parameter retrieval) 
    with ML, CV and high-performance computing techniques that answer `what questions', to the more ambitious goals of explainable AI, causal relations and ontologies that answer `what if' questions, and finally to communicate decisions, which involves ethical issues, trust and interaction with the user. 
}
    \label{fig:scheme}
\end{figure*}

More specifically, this article provides a comprehensive overview of the essential scientific tools and approaches that inform and support the transition from raw EO data to usable EO-based information. 
The promises, as well as the current challenges of these developments, are highlighted under dedicated sections. 
Specifically, we cover the impact of 
(i) computer vision and machine learning; 
(ii) advanced processing and computing; 
(iii) knowledge-based AI; 
(iv) explainable AI and causal inference;  
(v) physics-aware models; 
(vi) user-centric approaches; 
and (vii) the much-needed discussion of ethical and societal issues related to the massive use of ML technologies in EO.
Figure \ref{fig:scheme} summarises the content and organisation of this perspective paper.

\clearpage
\part{Modelling --- Machine learning, computer vision and processing}

% Section : computer vision
\input{./sections/2_cv_ml}

% Section : Adv. processing and computing
\input{./sections/9_computing}

\clearpage
\part{Understanding --- Physics-machine learning interplay, causality and ontologies}

\begin{figure*}[t]
    \centering
    \includegraphics[width=15cm]{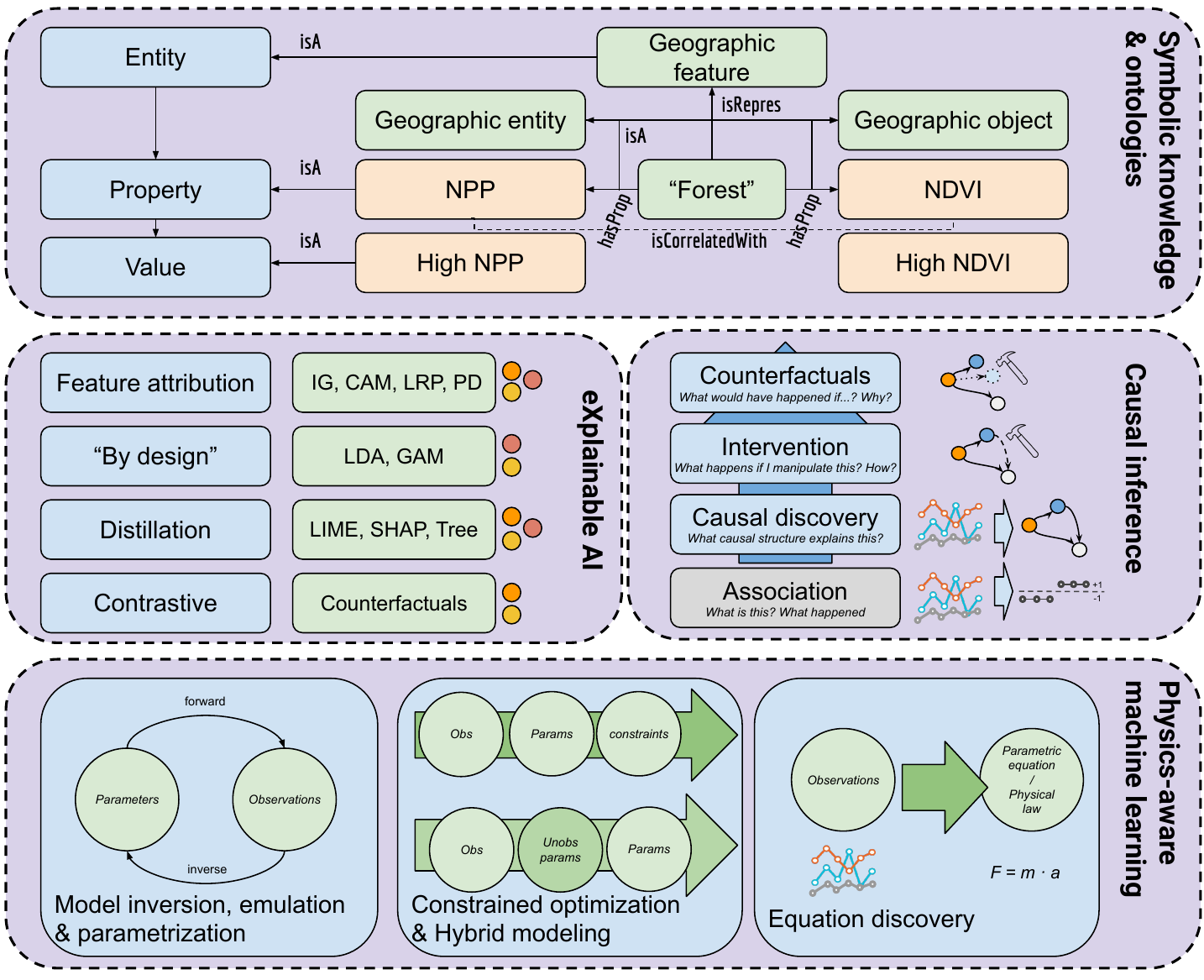}
    \caption{\small Methodologies for understanding complex systems such as the Earth. \textbf{Top:} Ontologies can represent both symbolic and numeric knowledge, reason only based on cognitive semantics and share knowledge on the interpretation of data, e.g. remote sensing images. Following \cite{arvor2019ontologies}, we can describe complex concepts such as ``forest'' as a particular `Entity' with some attributes or properties (e.g. NPP and NDVI levels) which can have associated values (e.g. high or low). This object can also be defined according to many intertwined properties and value levels. \textbf{Middle:} Many XAI methods are available to obtain explanations from ML models, typically categorised in feature attribution methods, explainable by design, distillation-based and contrastive \cite{ras2022explainable}, which fall under post-hoc methods (indicated by the orange dot), local (indicated by the yellow dot) or model-agnostic (indicated by the red dot). However, XAI only explains what the ML model learned, which can be correct for wrong reasons. Causality advances in the interpretability \cite{rudin2019stop,Pearl2000}, and proposes steps beyond the mere association rules developed in conventional ML -that can only answer `what' questions- by advancing in discovering causal structures in the data, answering `how' (intervention analysis) and `why' (counterfactuals) queries. \textbf{Bottom:} ML and domain knowledge can interact in many ways \cite{camps2018physics}: from inverting or emulating complex codes (e.g. RTM or climate models) to the estimation of parameters constrained by physics-based losses or ML-model coupling in hybrid modelling that allows estimating unobserved parameters \cite{Reichstein19nat}; and to the direct discovery of fully interpretable equations and laws from data \cite{CampsValls23physcausaldiscovery}.}
    \label{fig:dibujitos2}
\end{figure*}

%Section : symbolic AI
\input{./sections/6_symbolic}

% Section : causality, explainable AI
\input{./sections/3_xai}

% Section : hybrid models, physics-based
\input{./sections/4_physics}

\clearpage
\part{Communicating --- Machine-user interaction, trustworthiness \& ethics}

% Section : User-centric
\input{./sections/8_usercentric}

% Section : EO and society
\input{./sections/7_eosociety}

\addcontentsline{toc}{section}{Conclusions}
\section*{Conclusions}

The field of AI for EO is rapidly advancing due to the increasing availability of large amounts of sensor data at various resolutions. This perspective paper has provided an overview of the essential scientific tools and approaches supporting the transition from raw EO data to usable information. Since its inception, computer science and signal processing have been necessary in this field. Integrating machine learning, deep learning and computer vision techniques has recently opened up numerous avenues for further, substantial advances in EO{, with potential major impacts in downstream applications such as agriculture, food security, or environmental health}.

Here, we have discussed current challenges that still prevent the field from reaching full maturity and highlighted promising directions for future research leveraging approaches from computer vision, explainable AI, physics-aware models, user-centric approaches, and advanced processing and computing. 
Additionally, we have emphasised the importance of carefully considering ethical issues related to the massive use of ML technologies in EO, which is further heightened in the context of large generative ML models for EO. 
This overview aims to inspire the EO community to achieve the necessary transformative advances in the field and unlock future technical achievements. These are necessary to fulfil the monitor capabilities of using Earth observation to take the pulse of Earth.

\bibliographystyle{abbrv}

  \newpage
  
\section*{Contributions statement}  
\begin{itemize}
\item DT coordinated the writing group.
\item Section contributions (in \textbf{bold}, lead writer): 
Introduction: \textbf{DT}; 
Computer vision and machine learning: \textbf{KS}, \textbf{BD}, DT, JvR, HH; 
Advanced processing and computing: \textbf{BD}, VM, BLS, RS, FDF; 
Knowledge-based AI: \textbf{SD}, JvR, HH; 
Explainable AI and causality: \textbf{GCV}; 
Physics-aware ML: \textbf{FDF}, \textbf{GCV}, KS; 
User-centric: \textbf{DT}, XZ, MD; 
Ethics: \textbf{MK}.
\end{itemize}
ESA has partially funded the authors to work on this manuscript through the ESA visiting professors program.

\end{document}

%% file: sections/2_cv_ml.tex
% !TEX root = ../main-GRSM.tex
\section{{Machine learning and computer vision for Earth observation}}
\label{sec:cv_eo}

A significant portion of EO data comes in the form of images, so computer vision (CV) is a natural tool for extracting information from the data. 
Many techniques utilized in CV are rooted in the field of machine learning, such as neural networks. 
The EO community has benefited from ML-based techniques in computer vision while providing fertile ground for the development of new methods. For instance, neural network architectures developed specifically for image analysis are now commonly used in remote sensing, and the quest for human-like machine perception has inspired new research directions in EO, like monocular 3D perception or geospatial visual question answering.
Conversely, some EO-specific challenges do not typically arise in CV studies and call for targeted research efforts.

\begin{figure*}[tb]
    \centering
    \includegraphics[width=\textwidth]{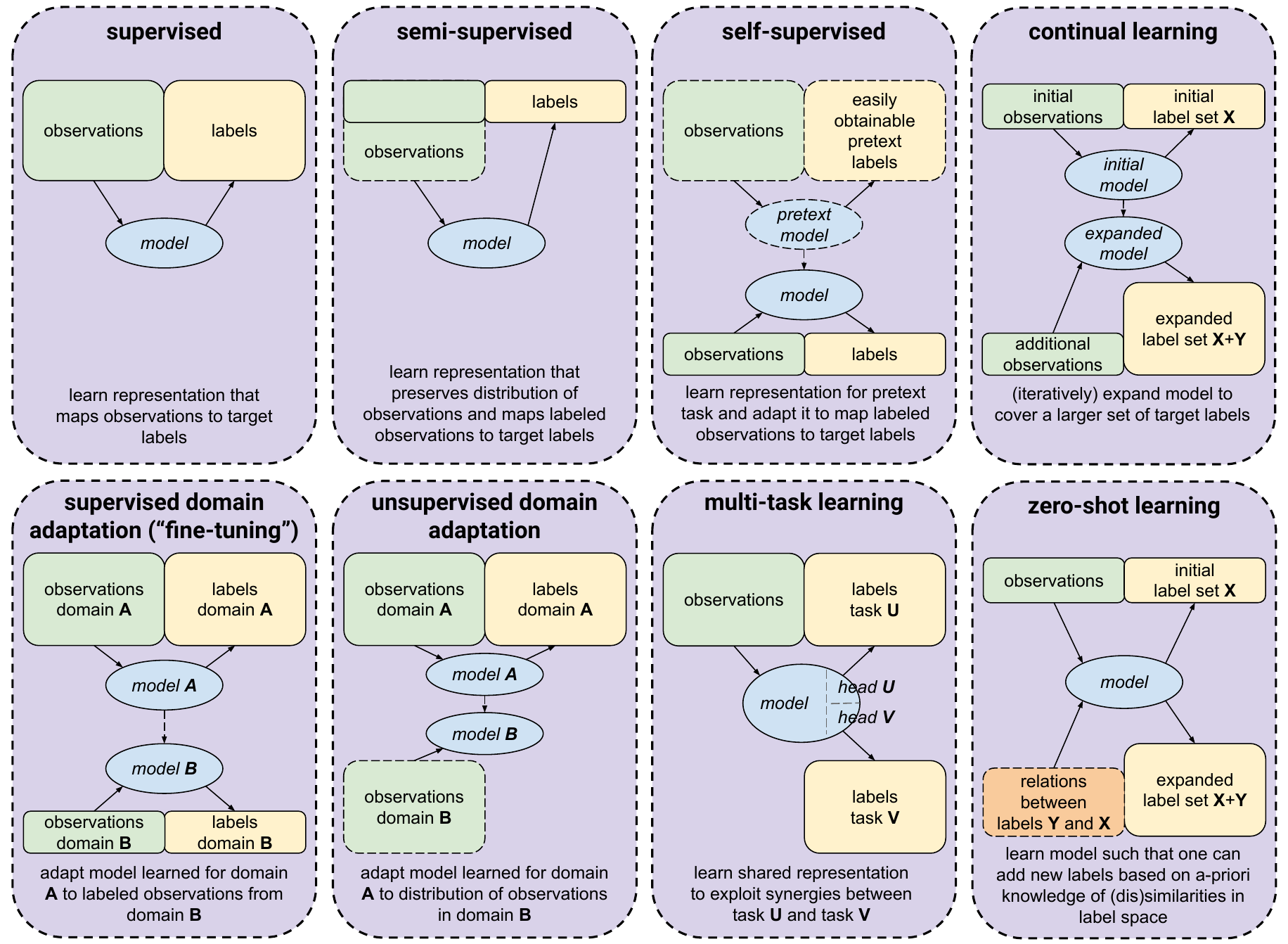}
    \caption{\small Different learning paradigms are relevant for AI applications in Earth observation, each with their own requirements and challenges. See text for details.}
    \label{fig:dibujitos_cvml}
\end{figure*}

\subsection{Specific challenges of Earth observation}

In remote sensing, data fusion~\cite{schmitt2016data} is routinely performed to jointly exploit multiple data sources (pan-sharpening, optical-SAR fusion, etc.). From the AI perspective, this is a form of representation learning where a joint latent representation shall be found for multiple data sources. It remains an open question how to optimally combine measurements from various sensors into a generic representation~\cite{scheibenreif2022contrastive}, while at the same time, we are confronted with an increasingly diverse range of data, including imaging sources, but also ground-based images~\cite{lefevre2017toward} or social media posts~\cite{jongman2015early}. Arguably, an ideal EO system promptly delivers downstream users the desired geospatial information using any suitable sensor data. This view suggests that EO services could benefit from user-centric abstractions (see Sec.~\ref{sec:usercentric}) that are tailored to specific information extraction tasks but flexibly draw from a pool of different inputs.

Another challenge is the synthesis between the AI data-driven paradigm and the rich physical knowledge about the Earth system~\cite{Cam20b}. 
Physical models excel at describing well-understood relations that can be condensed into explicit, low-dimensional equations.
On the other hand, the strength of statistical AI is to construct predictive models for phenomena where the mechanistic relations are poorly understood or intractable, by fitting generic, high-dimensional functions to large data samples.
For many quantities of interest in EO (hydrology, vegetation dynamics, etc.) physical models exist. However extracting them from data is a perception task, so one can expect a hybrid approach to be beneficial (see Sec.~\ref{sec:physics}). Not only would the results be consistent and explainable in terms of the underlying processes, but one would also need substantially less training data, yet still be more immune against overfitting.

A specific challenge in EO is the available reference data. Modern machine learning relies on large volumes of training data with labels (``ground truth'' prediction targets) that one might not have for EO, for instance, if collecting them involves fieldwork. For spatially explicit computer vision tasks it is often also assumed that the reference comes in the form of dense 2D label maps and that the labels show a balanced, representative sample. EO reference data is often not as ``analysis-ready''.
First, the reference may be imperfectly aligned to the images in terms of both location and scale, and it may refer to semantics not visible in the data, like land parcels. 
Second, ground truth may come as sparse, point-wise observations without surrounding context, making it harder to learn spatial patterns~\cite{lang2023nee}.
Third, the data distributions are often extremely unbalanced with long tails~\cite{waldner2019needle}, where rare, under-represented cases may be important (e.g. very high vegetation, uncommon crop diseases, weather extremes, etc.).

\subsection{General problem settings relevant for Earth observation}
\begin{figure}
    \centering
    \includegraphics[width=\columnwidth]{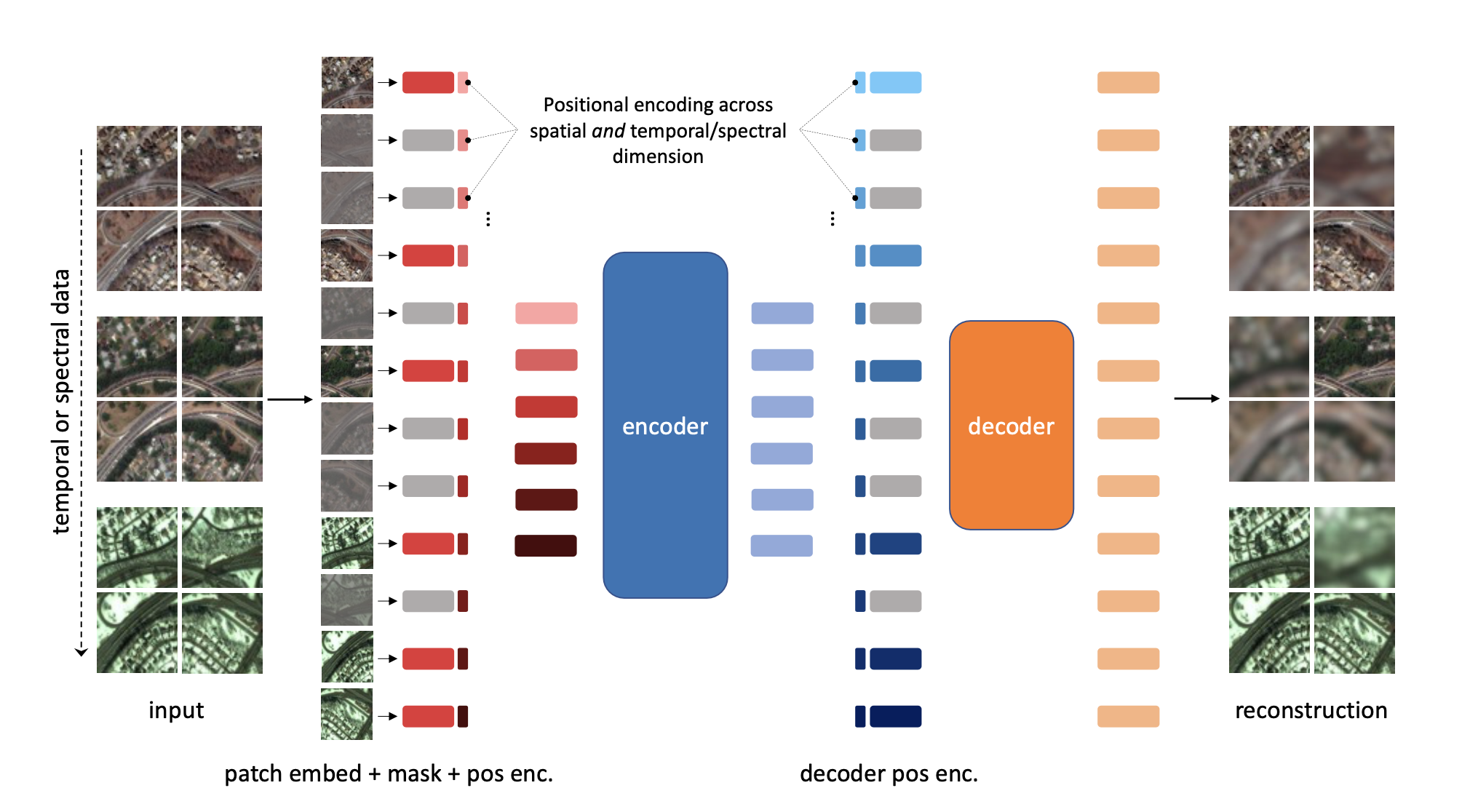}
    \caption{{Pre-trained transformer models for Earth observation are typically trained via the masked auto-encoder (MAE) principle. Image from~\cite{satmae2022} (\textcopyright 2022, BY-SA). }
    }
    \label{fig:satmae}
\end{figure}

Several problem settings from the general machine learning domain also emerge in the context of CV and EO. Figure~\ref{fig:dibujitos_cvml} shows several of these and how they relate to the more standard supervised and semi-supervised learning setting.
Arguably a main frontier of present-day AI is the step from supervised to unsupervised {(learning with no lables)} and \emph{self-supervised learning} (SSL{, learning in a supervised way with no labels by crafting a `pretext' learning task from the data itself -- e.g. mask part of the image and predict the masked part, predict the season the image was acquired in, etc.}). 
The current generation of computer vision systems relies primarily on supervised learning from large volumes of \emph{labelled} training data. 
Biological vision~\cite{fei2006one}, as well as information-theoretic considerations~\cite{hinton1999unsupervised}, suggest that visual perception can be learned with fewer annotations by abstracting inherent structure in the (unlabeled) data into a representation. 
While progress has been made, especially for language models~\cite{brown2020language}, unlabeled data are still not utilized to their full potential. 
There is a clear opportunity for future research, as in EO applications there are typically virtually unlimited quantities of unlabeled data available. When combined with advances in self-supervised learning {(e.g., learning via data augmentations~\cite{chen2020simple,chen2021exploring}, or by reconstructing masked image portions~\cite{he2022masked})} and large-scale pre-training, the abundance of unlabeled data is paving the way for foundational EO models that can be used across a wide range of retrieval tasks {(Figure~\ref{fig:satmae})}; 
potentially also including language or other information sources beyond EO sensor data~\cite{tseng2023lightweight,irvin2023usat,sun2022ringmo}.
{Both the access to large quantities of diverse, but unlabeled data and the synergies offered by multi-modality are in line with recent trends towards generative visual AI~\cite{ho2020denoising,rombach2022high} and foundational vision models~\cite{bachmann2022multimae,kirillov2023segment,jakubik2023prithvi}.} 

Due to the high amount of data that is being gathered (e.g., by satellites), \emph{continuously learning} new information (e.g. new LULC classes, sudden/abrupt changes) from a stream of recently acquired EO data is essential. 
This requires updating the model parameters by considering the recently acquired training images. 
On the one hand, re-training the entire ML model is impractical for large-scale EO applications due to computational and storage limits. 
On the other hand, updating the model parameters based on only recently available training samples may degrade the model's ability to characterize the previously seen data, an issue known as catastrophic forgetting~\cite{10.5555/3326943.3327027}. 
Few studies have addressed continual (deep) learning~\cite{8898615, 9184999, lenczner2022weaklysupervised,Sri19}{-- models that can make evolve their learning problem (e.g. the classes) without forgetting the one used for the original training --} and meta-learning~\cite{russwurm2022meteor} {-- models that instead of focusing on a single problem learn how to efficiently fine tune into new problems with minimal supervision --} for RS. 
Their primary goal is to overcome the forgetting of already learned LULC classes and leverage knowledge learned during earlier training rounds to obtain better performance or faster training for new LULC classes, which are considered new concepts. 
However, these methods require an accurate set of labelled data for the new classes, which may not be available in large-scale EO applications. 
Additionally, current methods assume that the concept boundaries are known. Such a setup is not well-suited for operational EO applications, where the concept boundaries are unknown a priori and need to be discovered. 
Further advances in continual learning are, therefore, likely to benefit the EO community.

However, many other learning paradigms can also be explored. 
\emph{{Semi-s}upervised domain adaptation} (or simply transfer learning) is a setting in which a neural network is trained on a large dataset and later adapted to a smaller dataset~\cite{chen2019closer}. 
This method turns out to be effective for cases where only limited labelled data of the concept is available, but many labelled data from a related concept are available.
\emph{Unsupervised domain adaptation}~\cite{sun2016deep} is a more complex problem setting in which there are no labels available for the target domain. 
Xu et al.~\cite{xu2022the} survey several remote sensing applications in this problem setting. 
There are several other prominent problem settings for which general ML methods have been developed that can be applied to EO applications, including but not limited to: few-shot learning~\cite{sung2018learning} to expand the output range to previously unseen cases with only a few examples; zero-shot learning~\cite{xian2017zero}, where the representation should have enough metric structure to handle output values outside the training range; and meta-learning to obtain flexible representations that can be transferred to new tasks with few data~\cite{brazdil2022metalearning,huisman2021survey}.

\subsection{AutoML for Earth Observation}
Many machine learning models used in EO are based on neural networks. 
To make them perform well, one must tune many architectural parameters, on top of the training hyperparameters~\cite{white2023neural}. 
{The goal of automated machine learning (AutoML) is to support the human in the data science loop by automating the design of a machine learning pipeline, including the choice of a suitable model, its architectural optimisation and the tuning of its hyperparameters}~\cite{hutter2019automated}. 
So far AutoML has largely been driven by benchmarks derived from standard CV applications (e.g., the NAS-Bench series~\cite{ying2019bench}) and has neglected EO-specific data such as BigEarthNet~\cite{sumbul2019bigearthnet}. 
Common characteristics of EO tasks are large amounts of observational data with relatively few high-quality labels and data distributions that substantially depend on the acquisition time and on the geographic location. To best support EO, AutoML must, therefore, go beyond the standard paradigm of selecting models and optimising their hyperparameters and should address more intricate topics like AutoML for heterogeneous spatiotemporal data.
This includes, for example, investigations on which type of search spaces would lead to good results within the constraints of the application~\cite{Hutter2014,perrone2019learning,Rijn2018}, or of the impact of acquiring more labelled data by modelling and analysing the learning curves of certain algorithms~\cite{mohr2022learning,mohr2023fast}. 

Moreover, recent advances in AutoML enable the configurations of models that are interpretable (c.f.\ Sec.~\ref{sec:xai}), robust against input perturbations~\cite{konig2024accelerating,silva2020opportunities} and that have well-calibrated uncertainty scores; properties that will also benefit EO applications.
To realise the potential of AutoML for EO, new EO-specific datasets are needed. 
While initiatives like BigEarthNet~\cite{sumbul2019bigearthnet} can be regarded as first steps towards that goal, no EO-related benchmarks have yet entered the mainstream to a point where researchers interested primarily in AutoML would adopt them and tackle their specific challenges~\cite{palacios2021automated,wasala2024autosr4eo}. 

\subsection{Beyond computer vision}

Finally, we argue that CV should not be viewed in isolation, but as a tool embedded in the larger context of Earth system monitoring and decision support.
From that perspective, an important question is how to best integrate AI systems into the EO infrastructure. {The variety of sensors used in EO is high -- arguably higher than in most other applications of CV. This creates a need for vision systems that can natively handle multi-sensor settings and exploit synergies between sensing modalities. To do so,} it may not be ideal to add CV capabilities to an EO program ex-post, instead they should inform the system design already in the planning phase. For instance, one could plan space missions with a view towards AI methods able to combine their respective observations. We postulate that systematically planned combinations, rather than the present, opportunistic practice~\cite{gavsparovic2018fusion,lang2023nee}, could greatly increase the impact of many satellite missions. Also for onboard AI processors deployed in space, e.g., to coordinate constellations or to preprocess data, it is obviously mandatory to include CV in the planning (see~Sec.~\ref{sec:onboard}).

Once AI is included in the system design, the question arises as to why it should be limited to perception and mapping. 
We speculate that, in the long run, CV will only be a component of much larger AI systems that go all the way to high-level reasoning and decision support.  Some AI researchers claim~\cite{chollet2019} that to that end one may have to integrate statistical AI with relational, symbolic AI (see Sec.~\ref{sec:kbai_eo}).

\begin{tcolorbox}[width=\columnwidth,title={Promising directions for future research}]    
   \begin{itemize}[leftmargin=10pt]
       \item[-] {Multi-modal EO models and cross-modal representation learning to move from sensor-specific to task-specific models }
       \item[-] {Self-supervised learning across varying combinations of tasks and/or spectral bands to build foundation models specialized in EO tasks} 
       \item[-] {Visual generative AI and conditional generative modelling for EO image analysis to tackle tasks as downscaling, sensor-to-sensor translation, data interpolation or simulation} 
       \item[-] {AutoML solutions specifically tailored towards EO problem settings}
       \item[-] {Domain adaptation, low-shot and continual learning to expand AI-based EO to new, data-scarce and/or dynamic applications}
       \item[-] {AI-friendly mission planning and EO/AI co-design to better integrate AI into space-based EO }
   \end{itemize}
   
\end{tcolorbox}

%% file: sections/9_computing.tex
% !TEX root = ../main.tex
\section{Advanced processing and computing}
\label{sec:computing}

This section describes the state-of-the-art and the most promising research directions in the efficient processing of very large and heterogeneous EO data.

\subsection{Earth observation ecosystems -- new system architectures}
\begin{figure}
\includegraphics[width=12 cm]{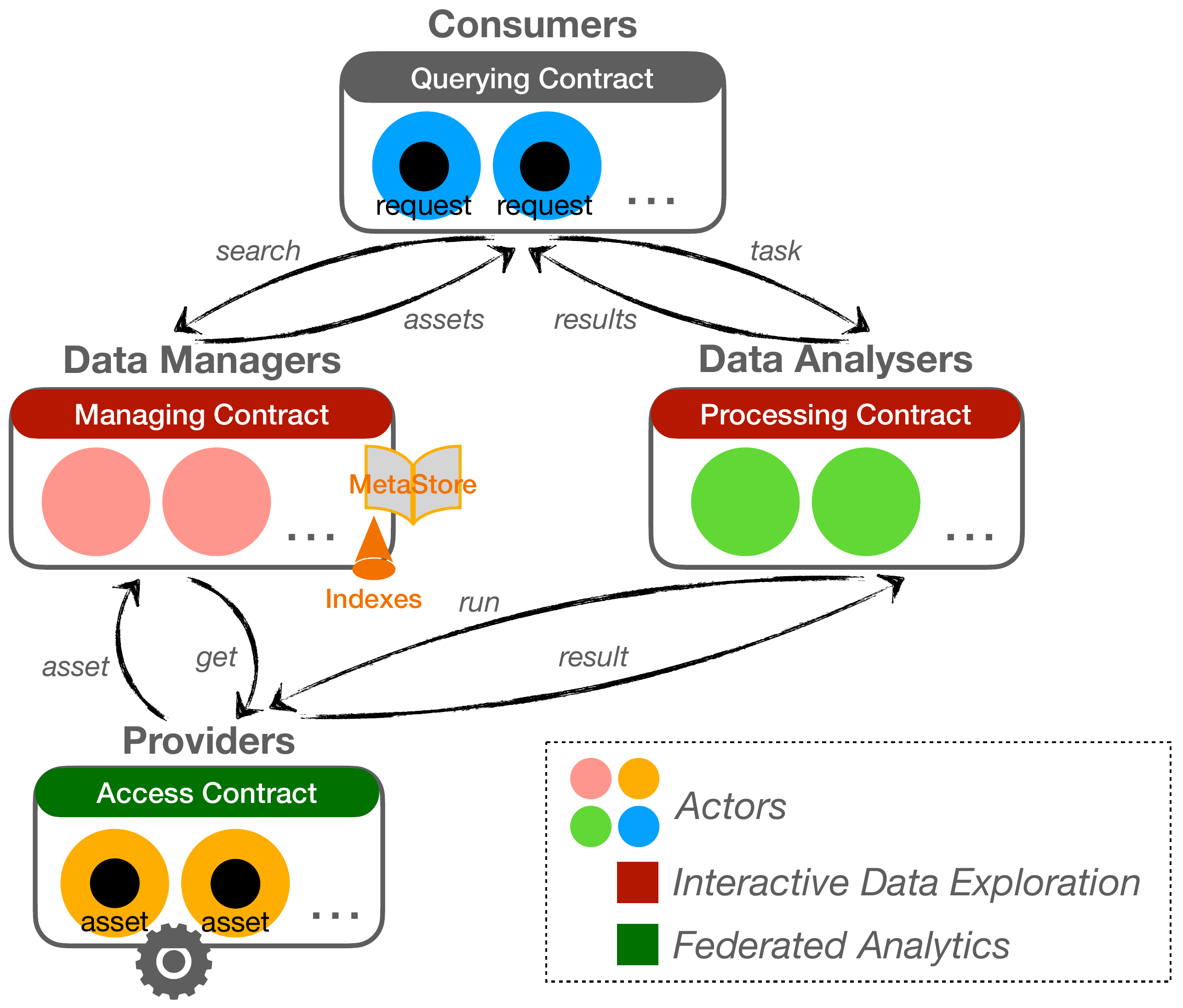}
\caption{{An Actor model-based EO ecosystem architecture. Image from~\cite{Wall2021} (\textcopyright 2021, BY-SA).}}
\label{fig:eoeco}
\end{figure}

Recently, different EO exploitation platforms --- such as the Google Earth Engine and Sentinel Hub -- have emerged to ease large-scale analyses by offering a certain level of processing and data abstraction~\cite{sudmanns_big_2020}.
According to our knowledge, all exploitation platforms rely on a heterogeneous set of technologies with varying sets of interfaces and data formats, making the federated use of these platforms difficult~\cite{Wall2021}. 
We believe that EO technology must be interoperable and accessible to everyone in a seamless manner. The AgoraEO project explores this direction toward a decentralized, open, and unified EO ecosystem, where one can share, find, compose, and efficiently execute cross-platform EO assets~\cite{Wall2021}. Figure~\ref{fig:eoeco} illustrates an architecture for an EO ecosystem based on the actor model, which provides the flexibility to shape an EO ecosystem to applications needs~\cite{Wall2021}.  
However, building EO ecosystems is still a big challenge. First, assets are highly heterogeneous, ranging from data archives, processing tools (e.g.~SNAP and GDAL), and ML algorithms to high-performance computing clusters and human expertise. This high heterogeneity renders asset management (storing, querying, and composition) an open problem~\cite{Traub2020}. Second, most EO analytics are exploratory, which requires fast query results. Third, data assets are naturally scattered across multiple sites and platforms, making AI-based analytics hard. It is necessary to devise a framework to support federated analytics.

\textbf{}

\subsection{Federated learning}
An increasing number of EO applications operate in highly distributed settings, especially with the emergence of EO ecosystems. Nowadays, it is not rare to find EO data highly scattered across multiple sites. 
However, standard ML algorithms require centralising the data before training, which is problematic in the EO settings due to its high volume. Additionally, centralising datasets might be impossible due to data constraints, such as privacy regulations or legal issues. 
Federated learning is a way to address this problem of associated governance, privacy concerns, and technical challenges~\cite{federatedlearning,federatedlearningblog, 10282873}. It aims to learn a local model on each distributed site (i.e., client) and aggregate the resulting local models to a shared global one~\cite{federatedlearning2017}. It reduces data transfer costs to a minimum, and models can generalise across EO datasets, leading to more accurate and less biased models. Applying federated learning to geo-distributed EO data is challenging because federated learning relies on the general assumption that the underlying data are independent and identically distributed (the i.i.d. assumption). This is rarely the case in operational EO data analysis scenarios. Furthermore, in RS, the data naturally tends to be remarkably heterogeneous – not only because of the variety of sensors, modalities, and characteristics of RS missions but even within the same product. Most of the existing federated learning algorithms have limited capabilities to deal with these conditions, leading to a sub-optimal performance~\cite{federatedlearningNon-IID}. {To overcome this issue, the first multi-modal federated learning framework in RS is introduced in~\cite{10282873}. This framework learns from decentralized multi-modal RS archives based on three modules: 1) multi-modal fusion module (which performs iterative model averaging to learn without accessing data on clients in the case that clients are associated with different data modalities); 2) feature whitening module (which aligns the representations learned among the different clients); and 3) mutual information maximization module (which maximizes the similarity of images from different modalities)}. Although this framework is found effective for multi-modal learning from decentralized and unshared data archives in RS, there is a need for the development of federated learning models that are associated to low local training and aggregation complexity, high learning efficiency and also low communication cost.

\subsection{AI interoperability}
The need for distributed or federated analytics and learning is increasing rapidly. More and more applications require building models in a distributed or federated fashion (i.e.~over multiple data sources).
Interoperability, in terms of data and model communication, as well as across different areas to achieve a common goal, e.g., building a global model, is critical in these applications. Without interoperability, achieving common goals efficiently and effectively is almost impossible. Primarily, we expect interoperability as a de facto standard in EO ecosystems, where multiple participants can join to contribute with their assets. This calls for standards representing input data (e.g.~satellite imagery) and intermediate data (e.g.~local models) to optimise multiple data sources and information models. Unfortunately, most data representations and systems are not interoperable at all. Examples in this direction are the Hub project by ActiveLoop, which aims at providing an AI database, i.e.~a common storage representation, for input data~\cite{activeloop}, the Open Neural Network eXchange (ONNX) format, an open-source standard for machine and deep learning~\cite{onnx}, or OpenML~\cite{Vanschorenetal2014}, an ecosystem for sharing ML datasets, algorithms and results. However, real interoperability is still far from reality in EO. AI interoperability must be solved if we expect EO ecosystems to succeed.

\subsection{On-board data processing}
\label{sec:onboard}
One of the most important goals of EO is to transfer satellite data processing from Earth (ground segment) to space (space segment) through the development of so-called onboard payload data processing. Indeed, quite often, the raw data generated by modern instruments is more than what can be transmitted to the ground. This makes using various signal processing and compression techniques necessary to reduce the amount of data transmitted, especially for small satellites. 
However, onboard processing requires several characteristics, such as flexibility, robustness and low computation burden. Moreover, the design of optimum schemes generally involves an appropriate combination of software and hardware solutions. For this reason, AI models represent a valid technology~\cite{Ziajaetal}. In this case, the highest processing demand is required during the training phase (which is operated on the ground segment), while less effort is necessary to yield the onboard products {with hardware prototyping that can be done in fast loops. We would like to note that for the onboard implementation, the computational complexity of deep neural networks can be reduced using various techniques, which further facilitate their application under the constraints of the space environment. As an example, a quantization technique that aims at reducing, with a negligible loss of accuracy, the precision of the numerical representation of the weights (e.g. from variables coded in single-precision floating-point format to those coded in signed integer format) is found effective for this purpose~\cite{gernigon2023low}.}
ESA has already launched a first demonstration mission named Phi-sat-1, characterized by onboard AI processing for cloud detection~\cite{murphy2021machine}. More challenging {and diverse} AI-based applications are being developed for implementation on the payload of the Phi-sat-2 mission~\cite{guerrisi2023artificial}. {The on-board reconfiguration of the AI components, either from the data perspective (data-centric) or from the module perspective (e.g. slimmable networks \cite{li2021dynamic}) represents one of the future trends of this technology and it is aligned with the more general goal of the next on-board imaging systems in space missions, which is to achieve partial or complete autonomy \cite{page2023developing}.}

\begin{tcolorbox}[width=\columnwidth,title={Promising directions for future research}]    
   \begin{itemize}
       \item[-]  {Development of decentralized, open and interoperable EO ecosystems}
       \item[-]  {Development of federated learning algorithms that are not only robust to training data heterogeneity across clients but also associated with low local training and aggregation complexity, high learning efficiency, and low communication cost}
       \item[-]  {Development of lightweight AI models for on-board processing of EO data} 
   \end{itemize}
   
\end{tcolorbox}

%% file: sections/6_symbolic.tex
% !TEX root = ../main-GRSM.tex

\section{Knowledge-based AI and Earth observation}

\label{sec:kbai_eo}

Knowledge-based AI includes knowledge representation, reasoning (incl. constraint satisfaction) and planning (incl. scheduling).
We discuss several opportunities for knowledge-based AI relevant to EO. {The top row of Figure~\ref{fig:dibujitos2} illustrates the main avenues in this direction}. 

\subsection{Decision modelling and decision support with Earth observation}

A prototypical example of the knowledge-based approach to AI is expert systems, designed to mimic the behaviour of human experts in narrow areas of expertise~\cite{Jackson1999}. They consist of a knowledge base (capturing human knowledge) and an inference engine (emulating human reasoning) and overall support the decision-making process. 
The spirit of expert systems has survived in its purest form at the intersection of AI and decision support and is exemplified by the DEX (Decision EXpert) approach~\cite{Bohanec2022}. DEX is a hierarchical multi-criteria decision modelling method, which represents decision preferences on alternatives described by qualitative variables in the form of decision rules, very similar to the IF-THEN rules of expert systems, and supports a variety of knowledge-explanation methods. 
DEX has been successfully and extensively used in various areas, with increasing applications concerning agriculture and its environmental impacts. One example of such an application is SoilNavigator, a field-scale decision support system for assessing and managing soil functions~\cite{Debeljaketal2019}. 
Due to the lack of measured data on soil functions, the SoilNavigator decision support system assesses all of them in a knowledge-based manner. 

The potential of EO data in decision-making is increasingly recognised. 
For example, the OECD~\cite{OECD} has prepared a document entitled ``EO for decision making'', which clearly states the importance of EO data for decision making, especially when combining it with other geospatial (e.g. ground data on air pollution), administrative and socio-economic data, in the context of assessing environmental risks and their impacts on humans and ecosystems. Combining knowledge-based approaches to decision support with EO data or downstream products derived by machine learning will unlock a whole new range of applications of AI that can help in achieving sustainable development goals~\cite{Vinuesaetal2020}.

 \subsection{Knowledge-based machine learning and Earth observation}

Most ML approaches today are data-driven and require large quantities of data to produce accurate and reliable models, which are typically not transparent and do not provide explanations. However, ML approaches that make use of formally represented domain knowledge, in addition to data, have a long tradition in the field, starting with explanation-based learning~\cite{Mitchelletall1986}. Inductive logic programming~\cite{LavracandDzeroski1994} uses logic programs to represent data (examples), domain knowledge (background knowledge) and learned models (hypotheses). This gives rise to a broader learning paradigm known as relational learning or relational data mining~\cite{DzeroskiandLavrac2001}. Combining logical representations with probability theory has led to the development of approaches that learn probabilistic models expressed in logic within the paradigm of statistical relational learning~\cite{GetoorandTaskar2007} and statistical relational AI~\cite{DeReadtetal2016}. 

The knowledge used in ML, in addition to data, can take different forms. Taxonomies are a form of domain knowledge that represents relations among concepts in the domain of study: in predictive modelling and classification, such taxonomies can represent hierarchical relations among the classes predicted. Typically, examples in this context have multiple labels from the taxonomy and the task at hand is known as hierarchical multi-label classification~\cite{SillaandFreitas2010}.
Land cover classification schemes~\cite{Yangetal2017} are typically hierarchical. However, most ML approaches to land cover classification ignore this and learn models for each land cover class separately. Multi-label classification of land cover~\cite{Stivaktakisetal2019,sumbul2020deep,mollenbrok2023deep} and classification with label hierarchies~\cite{Tui10b,turkoglu2021crop} is attracting more and more attention. 

Other forms of knowledge that ML approaches can use include constraints, as considered in constraint-based data mining~\cite{Dzeroskietal2010}. Rules can be hard or soft and can (in predictive modelling) impose preferences on the predictive power, form (language constraints), or size of the learned models. Constraints on clustering can, for example, specify that specific data points should belong to the same cluster. 

Last but not least, explanation and the use of existing domain knowledge play a prominent role in computational scientific discovery~\cite{DzeroskiandTodorovski2007}, which includes the topic of automated modelling of dynamical systems in science and engineering. Approaches such as process-based modelling, exemplified by the tool ProBMoT~\cite{Simidjievskietal2020}, integrate data-driven and knowledge-driven approaches into modelling, where ontological domain knowledge in the form of templates can be used as model parts. 

\subsection{Ontologies for open science in machine learning and Earth observation}

Ontologies formally represent knowledge about a domain of interest, focusing
on entities in the field and the relations between them. They mainly comprise
classes (or concepts), properties (or attributes), instances (or class members) and
relationships. Ontologies are crucial in providing controlled vocabularies for semantic annotation of datasets, which allows datasets to be FAIR, i.e. findable, accessible, interoperable and reusable~\cite{Wilkinsonetal2016}. With the rapidly increasing amounts of EO data, datasets must be FAIR according to the aforementioned criteria to maximise their added value. 

Currently, EO data are primarily managed through spatial data infrastructures, which facilitate the exchange, sharing, use and integration of geospatial data based on interoperability standards. Spatial data infrastructures follow the FAIR principles in some aspects (e.g. have registered or indexed meta-data in a searchable resource and are also accessible, even when the data are no longer available). However, they are not yet entirely FAIR-compliant~\cite{Giulianietal2021}, mainly due to the lack of ontologies for EO data that would provide vocabularies for meta-data that follow FAIR principles. 
We expect significant further development of ontologies for EO, ML, and other areas of AI. These should be connected to commonly accepted upper-level ontologies, facilitating their interoperability. Linking the ontologies for EO and AI will significantly facilitate open science at the intersection of EO and AI. {As a recent example, \cite{grujdin2023self} proposed a neuro-symbolic paradigm for a self-supervised ontology learning for natural hazard identification and monitoring.}

\begin{tcolorbox}[width=\columnwidth,title={Promising directions for future research}]    
   {\begin{itemize}
       \item[-] Utilize Earth observation data to derive better decision support systems
       \item[-] Utilize domain knowledge to be included as input in machine learning systems
       \item[-] Development of extensive ontologies for Earth observation
   \end{itemize} }
\end{tcolorbox}  

%% file: sections/3_xai.tex
% !TEX root = ../main-GRSM.tex
\section{Explainable AI and causal inference}
\label{sec:xai}
Experimentation and observational data analysis form the foundation for accurately modelling and understanding physical phenomena \cite{CampsValls23physcausaldiscovery}. However, model fitting alone is often inadequate, and there is a desire to comprehend and characterize the system's behaviour. The scientific consistency, reliability, and explainability of results generated with AI models are crucial when dealing with complex systems such as the Earth or its climate systems~\cite{Reichstein19nat,Tui20grsm,Runge19natcom}. Developing robust models that can be visualized, queried, and interpreted is essential. Achieving transparency, interpretability, and explainability in ML models for geosciences and RS is key for several reasons: 1)~to ensure the model's performance consistency, generalizability, and robustness, 2)~to enhance understanding of the Earth's system and 3)~to foster broader adoption and confidence among domain scientists. However, model interpretability is challenging because the model assumes a specific causal relation (inputs cause outputs) that may not be correct or complete. Retrieving causal relations from observations and quantifying the effects of certain interventions has been a longstanding human goal and remains a key objective when employing AI approaches, particularly ML techniques, for scientific inquiries \cite{Runge23causalreview}. {This section will explore recent advances in explainable AI and causal inference for EO. The second row of Figure~\ref{fig:dibujitos2} summarizes the main concepts presented.}

\subsection{Explainable AI}
Due to their size and complexity, many successful neural network models pose a challenge in comprehending or studying their inner workings at a level of abstraction that provides insight into their predictions. Various explainable AI (XAI) methods exist for different learning paradigms~\cite{SamArXiv20,roscher2020explainable}. These methods encompass {\em post-hoc interpretability}, employing advanced feature ranking techniques~\cite{Ola19xai,Johnson20sakame} or analyzing activation maps within the learned network (e.g., GradCam or layer-wise relevance propagation), as well as {\em interpretability by design}, involving networks explicitly designed to be explainable~\cite{Marcos2019}. These techniques can generate spatially explicit and temporally resolved explanations of what AI models have learned, emphasizing specific spatial and temporal scales~\cite{Johnson20sakame}.
 
``Post-hoc'' methods, including backpropagation-based or feature attribution techniques like layer-wise relevance propagation, DeepLift, and integrated gradients~\cite{roscher2020explainable,SamArXiv20}, also encompass input feature ablation, occlusion, and perturbation-based approaches such as model reliance and Shapley value sampling. These methods address the challenge of meaningfully relating a complex neural architecture's output or intermediate activations to its input features. 
Attention mechanisms and transformers, which have shown promise in Earth sciences~\cite{russwurm2020self}, provide ante-hoc explanations by assigning coefficients to different spatio-temporal locations, indicating their importance when the model generates an output. These coefficients form saliency maps, facilitating interpretation. In contrast to traditional recurrent neural networks, attention mechanisms allow insight into long-range interactions in space-time series. 

``Explainable by design'' methods embed interpretability into the model. Examples include models based on semantic bottlenecks, where intermediate processing pipelines (e.g., a classifier of concepts relevant to the downstream task) are learned concurrently with the final task. Recent literature in EO proposes such models to understand subjective appreciation of landscapes~\cite{levering2021relation}, align forest type classifiers with foresters' decision-making processes~\cite{nguyen2022mapping}, or unsupervised exploration of what defines a wild landscape~\cite{stomberg2021jungle}.

\subsection{Causal inference}

Despite the impressive predictive capabilities of current machine and deep learning methods, there is limited actual learning, as these algorithms excel in fitting complex functions without providing a clear understanding of underlying causal relations~\cite{Pearl2000,Runge19natcom,Diaz21rccm,Runge23causalreview,CampsValls23physcausaldiscovery}. Analyzing Earth's complex systems with multivariate and non-gridded datasets faces missing data, non-linearities, and non-stationarities. Variables and processes are often coupled in space and time, with (tele)connections being long-range, discontinuous, and varying in strength. Addressing this issue is crucial for identifying predictors, developing robust models, and ensuring correct answers for the right reasons.

Causal inference aims to discover and explain a system's causal structure based on data, models, and assumptions~\cite{Pearl2000,Peters18,Runge23causalreview}. It addresses fundamental Earth science problems: 1)~hypothesis testing, 2)~discovering latent factors beyond spurious correlations, and 3)~understanding systems from observational data and assumptions. True understanding involves knowing the entire causal chain, enabling predictions of consequences and analysis of interventions and counterfactuals. Three main fields emerge as relevant for modelling and understanding processes in the Earth system.

\paragraph{Interventional Analysis}
Performing interventions on simulations, though challenging, enhances the efficiency and robustness of causal discovery. The choice of interventions requires careful consideration, respecting the spatio-temporal trajectory of the system. Physical and causal sense is crucial, ensuring a genuine distributional shift in the system's dynamics \cite{Runge23causalreview}.

\paragraph{Counterfactuals Analysis}
Counterfactual analysis imagines alternative outcomes based on different actions. Attributing causes of extreme events to climate change involves counterfactual thought experiments. Using algorithms like MACE, DACE, and NSGA-II, counterfactual modules facilitate model interpretation in human-like language, fostering explicit user-machine interaction \cite{Runge23causalreview}.

\paragraph{Causal Discovery}
Observational causal discovery extracts cause-effect relationships from multivariate datasets, moving beyond correlation. Various methods allow learning causal relations even without explicit time involvement, including probabilistic graphical models, structure learning, dynamical systems, Granger-causality, instrumental variable paradigms, and additive noise models. Understanding complex phenomena related to climate change requires discovering teleconnection patterns, where recent deep learning approaches show promise~\cite{varandolearning}.

\begin{tcolorbox}[width=\columnwidth,title={Promising directions for future research}]    
   \begin{itemize}
       \item[-] {Causal machine learning promises to integrate ML models with causal inference to better understand and predict complex earth systems dynamics observed through remote sensing}
       \item[-] {Exploring spatiotemporal causal models can lead to significant breakthroughs in discovering the impacts (effects) of environmental factors on global change}
       \item[-] {Investigating counterfactual scenarios through remote sensing and climate data can provide insights into the potential impacts of different interventions e.g., on socio-economic factors, aiding policymakers in making informed environmental decisions}
   \end{itemize}
   
\end{tcolorbox}  

%% file: sections/4_physics.tex
% !TEX root = ../main-GRSM.tex
%\newpage
\section{Physics-aware machine learning}
\label{sec:physics}

Machine learning has been widely and openly criticised because of the difficulty of interpreting models and the lack of physical consistency. It may happen that models are so big and overparameterised that accuracy compromises fulfilling physical laws, such as mass or energy conservation. Losing contact with physics should always be avoided if we want to design robust and reliable (ML) models. This disconnection has implications for the consistency of products and parameter estimation in data assimilation schemes and forecasting tools and in understanding the underlying processes from Earth's observational data.

Historically, physical modelling and ML have been treated as two fields under very different scientific paradigms (theory-driven versus data-driven). However, integrating domain knowledge and physical consistency has been suggested as a principled way to provide solid theoretical constraints on top of the observational ones~\cite{Reichstein19nat}. The synergy between the two approaches has gained attention by either redesigning the model's architecture, augmenting the training dataset with simulations, or including physical constraints in the cost function to be minimised~\cite{svendsen17jgp,Reichstein19nat}. 
Integrating physics in ML models may improve performance and generalisation and, more importantly, consistency and credibility. 
The hybridisation has an interesting regularisation effect, given that the physics limits the parameter space to search and thus discards implausible models. 
Therefore, physics-aware ML models better combat overfitting, typically become simpler (sparser), and require less training data to achieve similar performance. 
Physics-aware ML thus leads to enhanced computational efficiency and constitutes a stepping stone towards achieving more interpretable {and robust} ML models~\cite{Samek2019,von2019informed}. {This section presents advances in physics-aware machine learning and its applications to the specificities of EO. The third row of Figure~\ref{fig:dibujitos2} summarizes the main concepts covered in the section. }

\subsection{Inverting physical models with machine learning}
Biogeophysical parameter estimation and retrieval with ML aim at learning a mapping from an observed satellite spectrum to an underlying biophysical parameter. Unfortunately, relying only on observational data (both spectra and in-situ measurement) to perform regression is expensive, time-consuming, and very often infeasible~\cite{camps2016survey,camps2019perspective}. This problem can be better posed by exploiting the wealth of simulated data available through physical models such as radiative transfer models (RTMs) and electromagnetic models (EMs). They solve what is called the \emph{forward problem}, that is, they generate physically meaningful observational data (e.g. spectra) from state parameters (e.g. leaf-level or canopy parameters). Once the simulations 
of a particular observed bio-geophysical scenario (e.g. a forest) are given, they provide the electromagnetic quantity that would be measured by the EO sensor (e.g. the back-scattering coefficients or reflectances).  
Using these models, we can generate as many input-output pairs as we need for training and testing our ML scheme to do the inversion, i.e. to estimate parameters from RS data~\cite{del2004neural,sellitto}. {Another example is the prediction of quad-pol signatures for single-polarized SAR data~\cite{datcu2023explainable}.} 
The use of an RTM/EM model can be beneficial from different points of view: (1)~they can help with data generation balancing by compensating for the lack of ground measurements; (2)~they can help in understanding the bounds within which the desired parameters can be retrieved less ambiguously; and (3)~they can help in the selection of the optimal inputs of the ML model.

\subsection{Hybrid machine learning: interactions of transfer models and ML}
ML algorithms and physics can be fully blended in several ways extending traditional approaches based on data assimilation. Still, simple constrained optimisation methods often work well in practice~\cite{kashinath2019physics,wu2018physics}. Indeed, by inferring the constraints to the input/output pairs based on physical models, ML models can obtain improved performance. 
Another option is to learn ML emulators combined with purely data-driven algorithms for model inversion~\cite{camps2018physics}. Finally, one can also consider a fully coupled neural network where layers that describe complicated and uncertain processes feed physics layers that encode known relations of intermediate feature representation with the target variables. The promise of such a hybrid modelling approach~\cite{camps2020advancing} is to combine the excellent predictive abilities of modern ML, fuelled by the ``unreasonable effectiveness of data''~\cite{halevy2009unreasonable}, with the causal consistency, explainability, and data efficiency of classical physical modelling. 

Physical models require setting parameters that are rarely derived from first-domain principles. ML models can learn such parameterisations. For example, instead of assigning vegetation parameters empirically to plant functional types in  Earth system models, one can learn these parameterisations from proxy covariates using ML, and thus achieve flexibility, adaptability, dynamics, and context-dependent properties~\cite{moreno2018methodology}. 

Emulating models in geosciences, climate sciences, and remote sensing is gaining popularity~\cite{camps2016survey,Reichstein19nat,camps2019perspective}. Emulators are ML models that mimic the forward physical models using a small yet representative data set of simulations. Once trained, emulators can provide fast-forward simulations, allowing improved model inversion and parameterisations. However, replacing physical models with ML models requires running expensive offline evaluations first, and alternatives exist that construct the model and choose the proper simulations iteratively~\cite{camps2018physics,Svendsen19amogape}. 

\subsection{Discovering equations from data} 

As in many fields of science and engineering, Earth system models describe processes with a set of differential equations encoding our prior belief about the dynamics and variable interactions. Learning ordinary and partial differential equations (ODE/PDEs) from stochastic observations is perhaps one of the most challenging problems nowadays in statistical learning. Several approaches exist for solving this problem, ranging from equation-free modelling~\cite{ye2015equation} and empirical dynamic modelling~\cite{Diaz21rccm} to automated inference of dynamics~\cite{daniels2015automated}. Imposing sparsity on equation discovery is a sensible assumption that allows us to identify the governing equations of the biosphere more sharply~\cite{brunton2016discovering}. Other criteria, such as compositionality, unification of theories, and interestingness, can help in learning laws and equations from data while trying to alleviate the {\em equifinality} (non-identifiability) issue, that is, finding many equally accurate models for the wrong reasons. For a comprehensive overview of equation discovery techniques with case studies in Earth and climate sciences, we refer the interested reader to~\cite{CampsValls23physcausaldiscovery}.

\begin{tcolorbox}[width=\columnwidth,title={Promising directions for future research}]    
   \begin{itemize}
       \item[-] {Using hybrid machine learning to emulate costly radiative transfer models (RTMs) to improve the speed and accuracy of EO data analysis, allowing more trustworthy and interpretable simulations}
       \item[-] {Hybrid approaches allow for more precise parameter estimation in Earth science models that enable more accurate predictions, calibrated uncertainties, and a better understanding of environmental processes}  
       \item[-] {Hybrid models yield sparser, more interpretable ML models that enhance transparency and usability in EO, making complex data more accessible and actionable}
   \end{itemize}
   
\end{tcolorbox}  

%% file: sections/8_usercentric.tex
% !TEX root = ../main-GRSM.tex
\section{User-centric Earth observation }
\label{sec:usercentric}

Making EO data accessible in catalogues is not enough. Improving users' experience must be high on the agenda too. Such experience is currently still not optimal, since EO data analysis and interpretation is still performed in a laborious way, by repeated cycles of trial and error, without reaching the desired degree of flexibility and robustness. Furthermore, the overwhelming amount of data puts strong limitations on the extent (geographic coverage within acceptable costs), depth (causal analysis, evaluation of the implications) and response time of human interpretation. As a consequence, a significant amount of EO data remains  unused~\cite{filchev_lachezar_2018_2475063}.
User-centric EO aims at enabling the interpretation of EO data content, associating it with other sources of information, understanding user inquiries in an interactive dialogue (thus refining the expression of needs), distil data content, and suggesting the most appropriate information items or the interpretation alternatives.

\subsection{Decreasing the user load with active learning}%active learning
\label{sec:usercentric:al}
The first form of interaction is to be found in the support of human operators creating the labels. Extracting labels directly from images without guidance is time-consuming and can be difficult to crowdsource. EO crowdsourcing may not be impossible, but is hampered by factors such as the lack of agreement about class definitions among communities, domains and even users. The label collection process can be optimized via human-machine interaction with active learning~\cite{tuia09active}. {The active learning process is iterative: at each iteration, the learning algorithm automatically chooses the most informative unlabeled samples for manual labelling from a human expert (supervisor) and the algorithm is retrained with the additional labelled samples. This way, the unnecessary and redundant labelling of samples that are not informative for the classifier is avoided, significantly reducing the labelling cost and time.} Questions of sample informativeness (uncertainty and diversity of samples) have been at the core of the debate. At the same time, equally important aspects such as accessibility costs~\cite{demir2013definition} and user's skills in labelling~\cite{tuia13learning} have been relatively neglected so far. Also, the question of the applicability of active learning methods to DL models is still an open debate, since single samples' usefulness is diluted by the batch-learning nature of deep neural networks~\cite{rodriguez2021mapping} or focusing on sample diversity and misclassifications of the current model~\cite{Kel20birds} are also being actively explored. A comprehensive and systematic overview of deep active learning methods is provided in~\cite{deepAL}, showing promising avenues primarily on improving sample selection strategies, optimising training methods and improving task-independent models.

\subsection{Language-based interaction}%VQA
\begin{figure*}[!t]
    \centering
    \includegraphics[width=16cm]{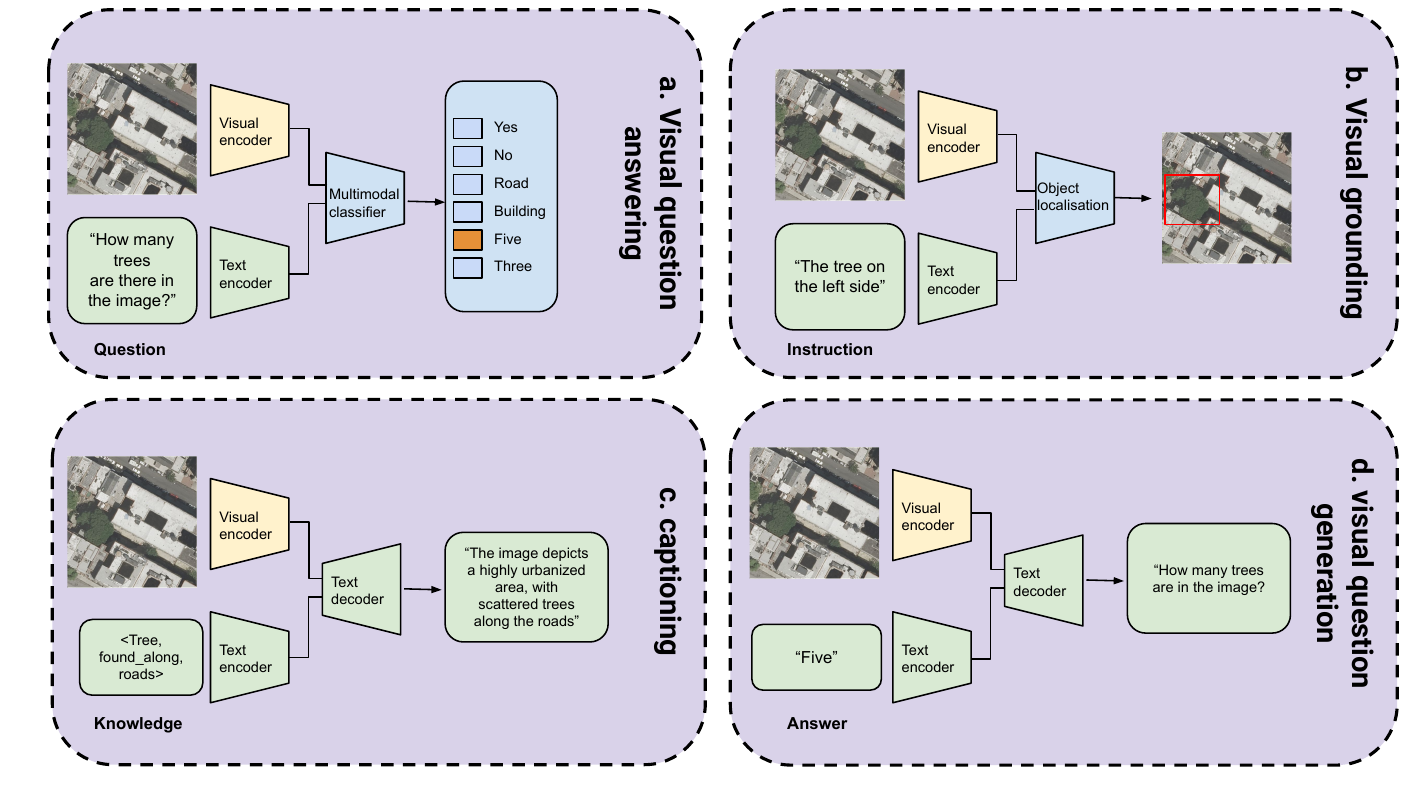}
    \caption{Main current advances at the interface of vision and language for remote sensing: a) visual question answering: answering a text question about an image, b) visual grounding: identifying objects in the image with text; c) captioning: describe the content of the image, possibly using some external knowledge; d) visual question generation: generate a question about the image that has the specified answer (could also use external knowledge). In the plots, yellow parts are visual encoders, green are text-based encoders/decoders, and blue are ML models for classification / object detection. Inspired by \cite{bashmal2023language}.}
    \label{fig:vlms}
\end{figure*}

\begin{figure*}[ht]
    \centering
\begin{minipage}[htb]{0.9\linewidth}
%    \centerline{
            \begin{tabular}{c|c|cccc}
            \centering
              \multirow{2}{*}{(a) {RGB}}  & \multirow{2}{*}{(b) {Prediction}}  & 
              \multicolumn{4}{c}{Text prompts}\\
&&              (c) {Roads}  & (d) {Agric. and lake} & (e) {Swimming}  & (f) {Squirrel}  \\
            \includegraphics[width=.152\linewidth]{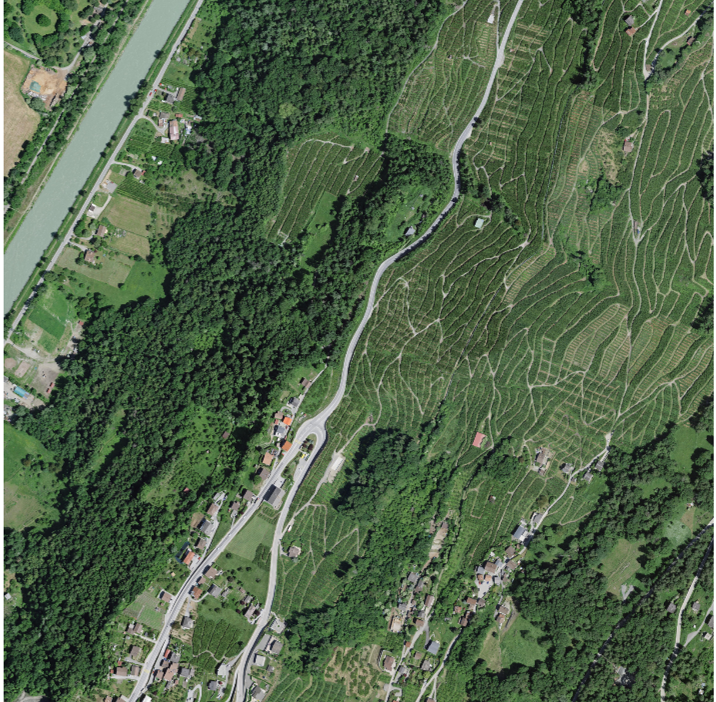} &
            \includegraphics[width=.15\linewidth]{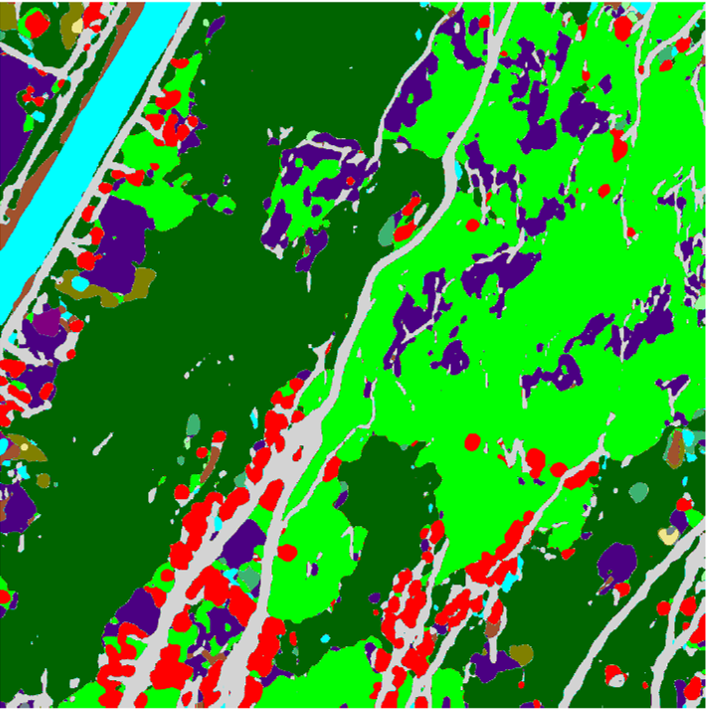} &
            \includegraphics[width=.152\linewidth]{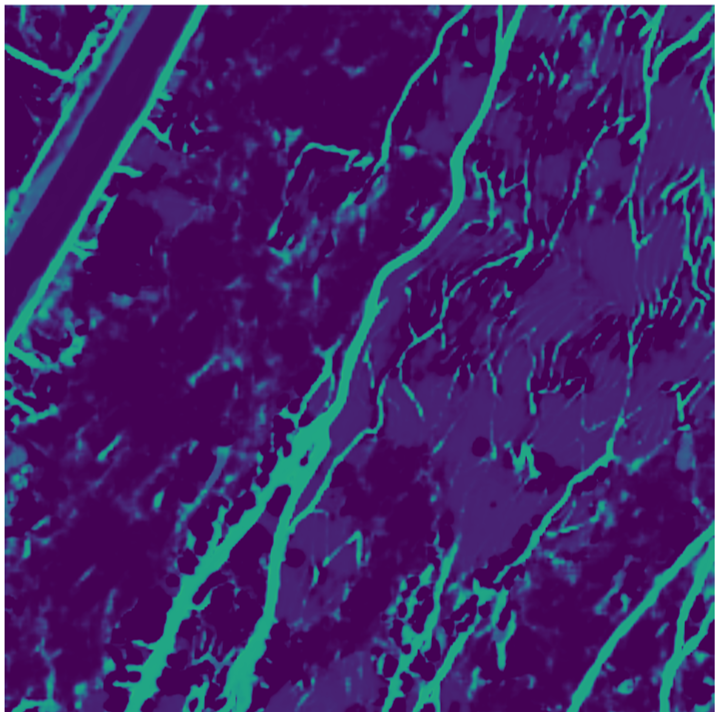} &
            \includegraphics[width=.15\linewidth]{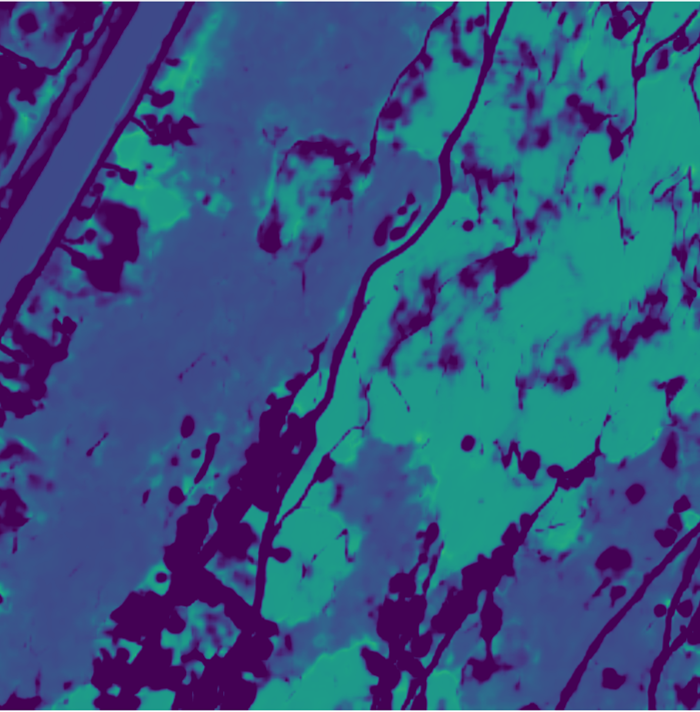} &
            \includegraphics[width=.15\linewidth]{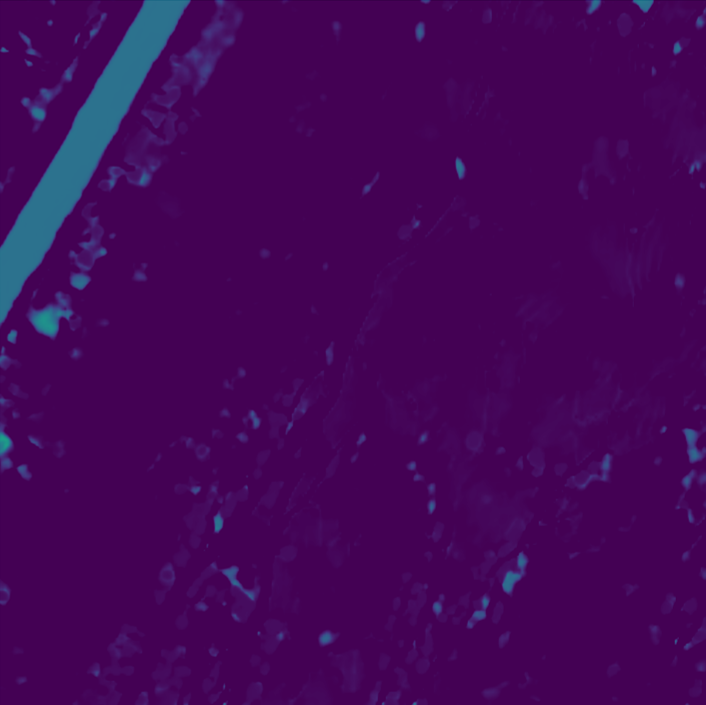} &
            \includegraphics[width=.15\linewidth]{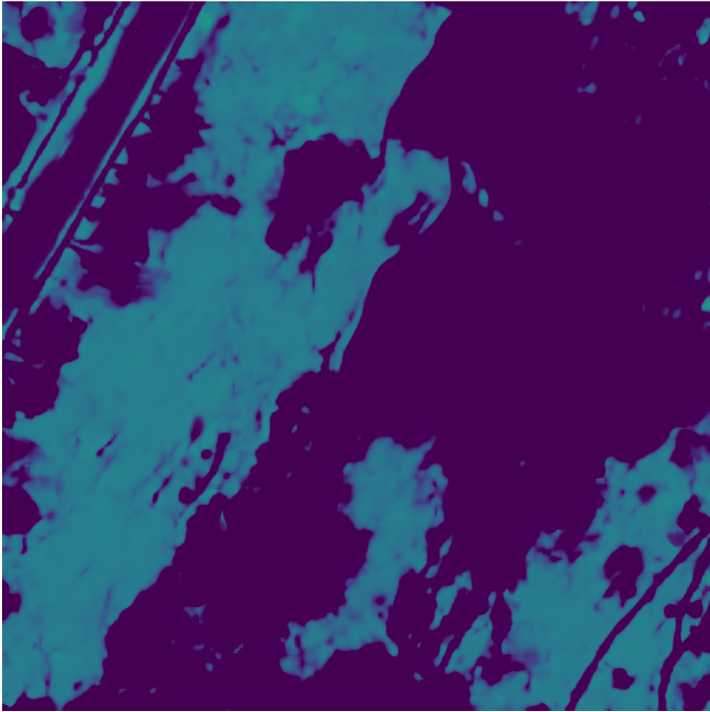}  \\
            \includegraphics[width=.152\linewidth]{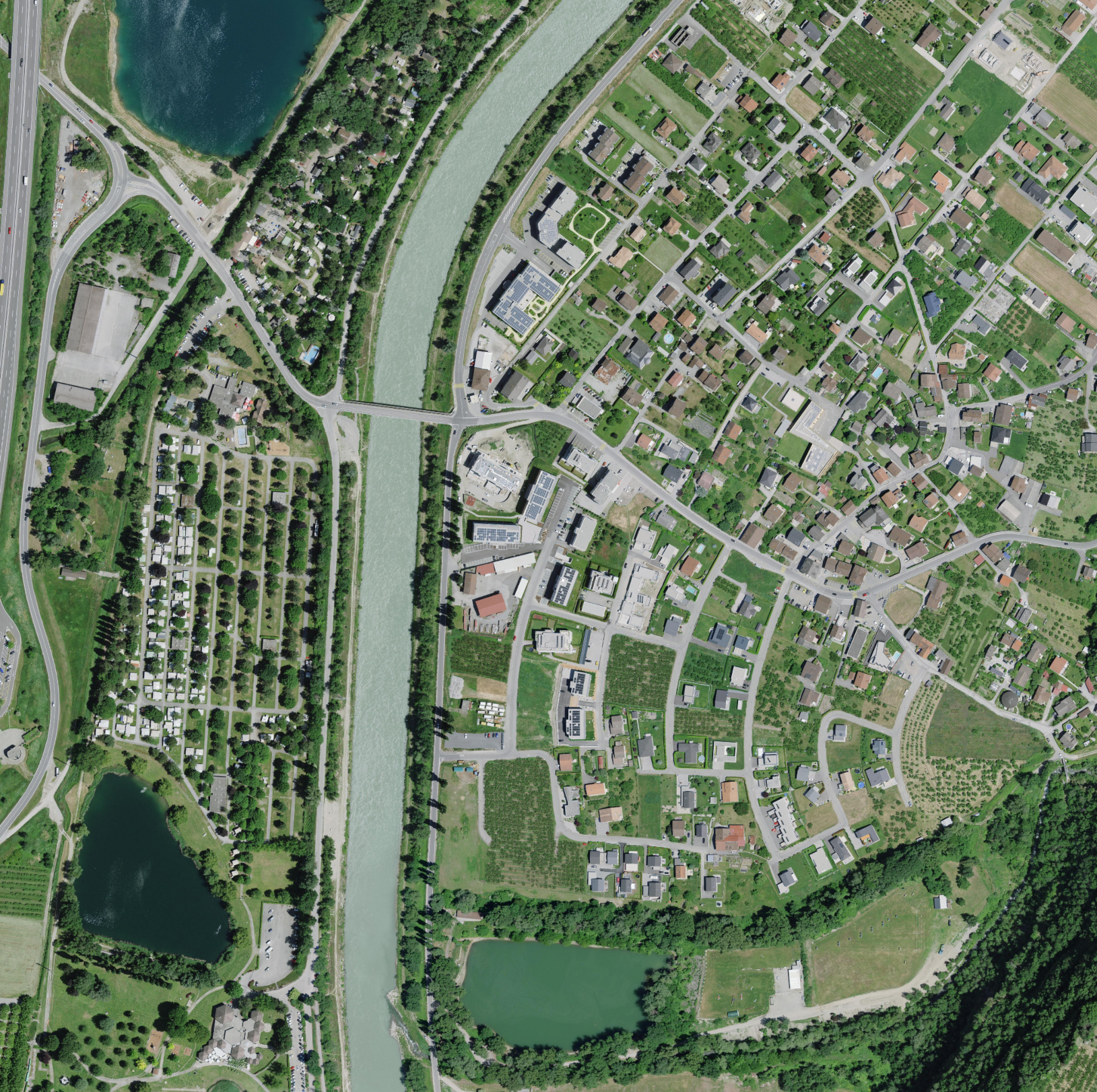} &
            \includegraphics[width=.15\linewidth]{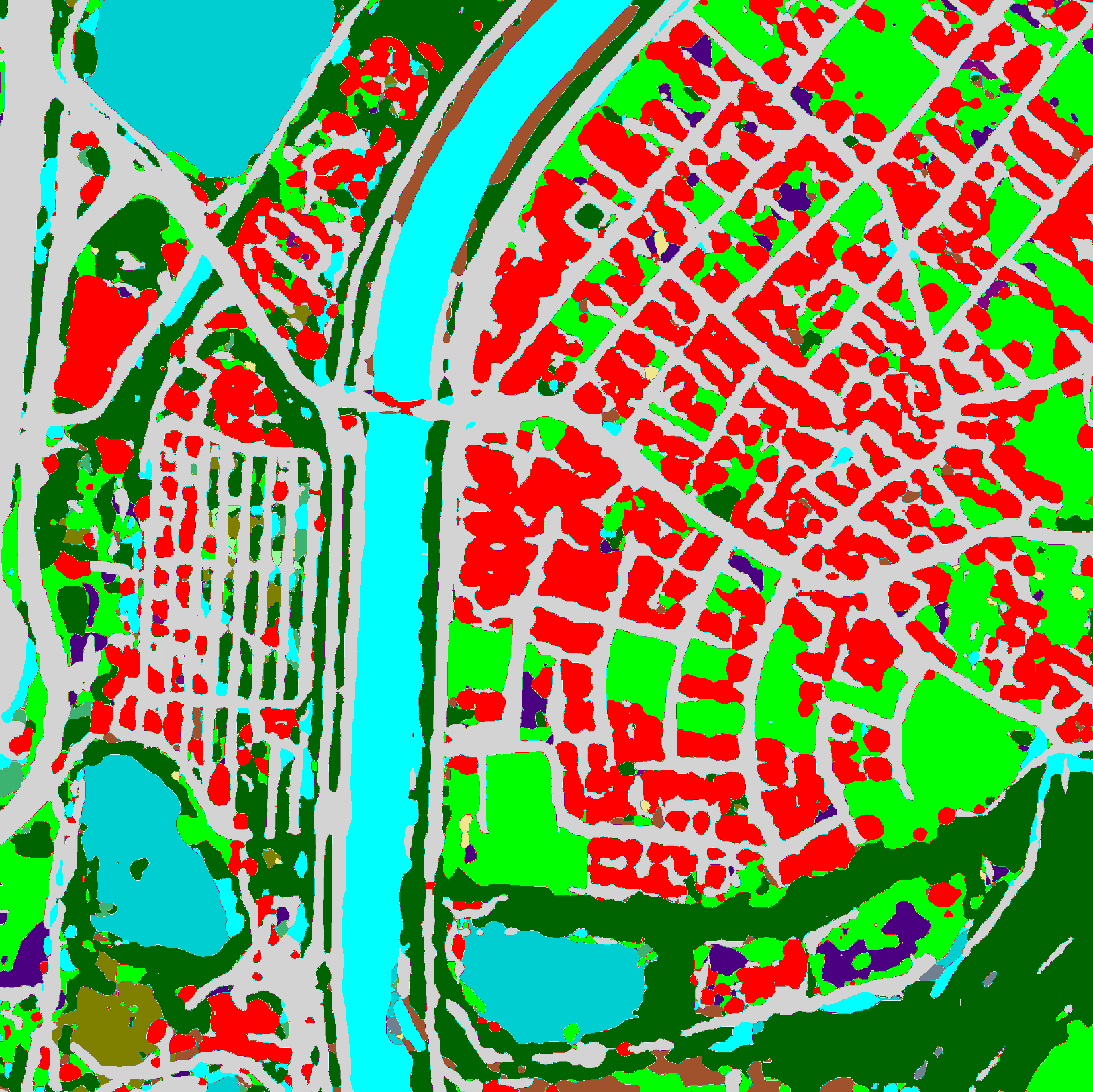} &
            \includegraphics[width=.152\linewidth]{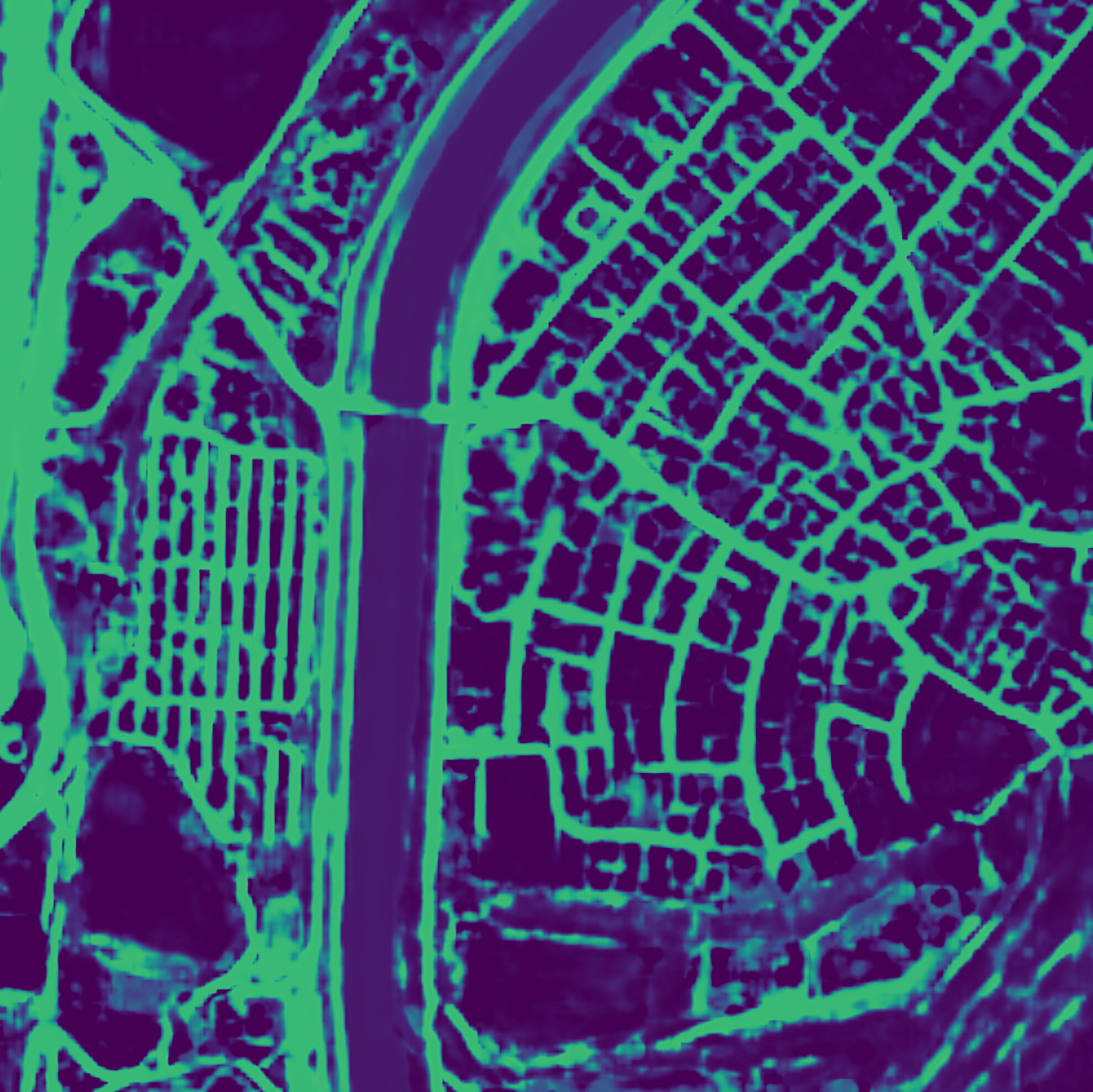} &           
            \includegraphics[width=.15\linewidth]{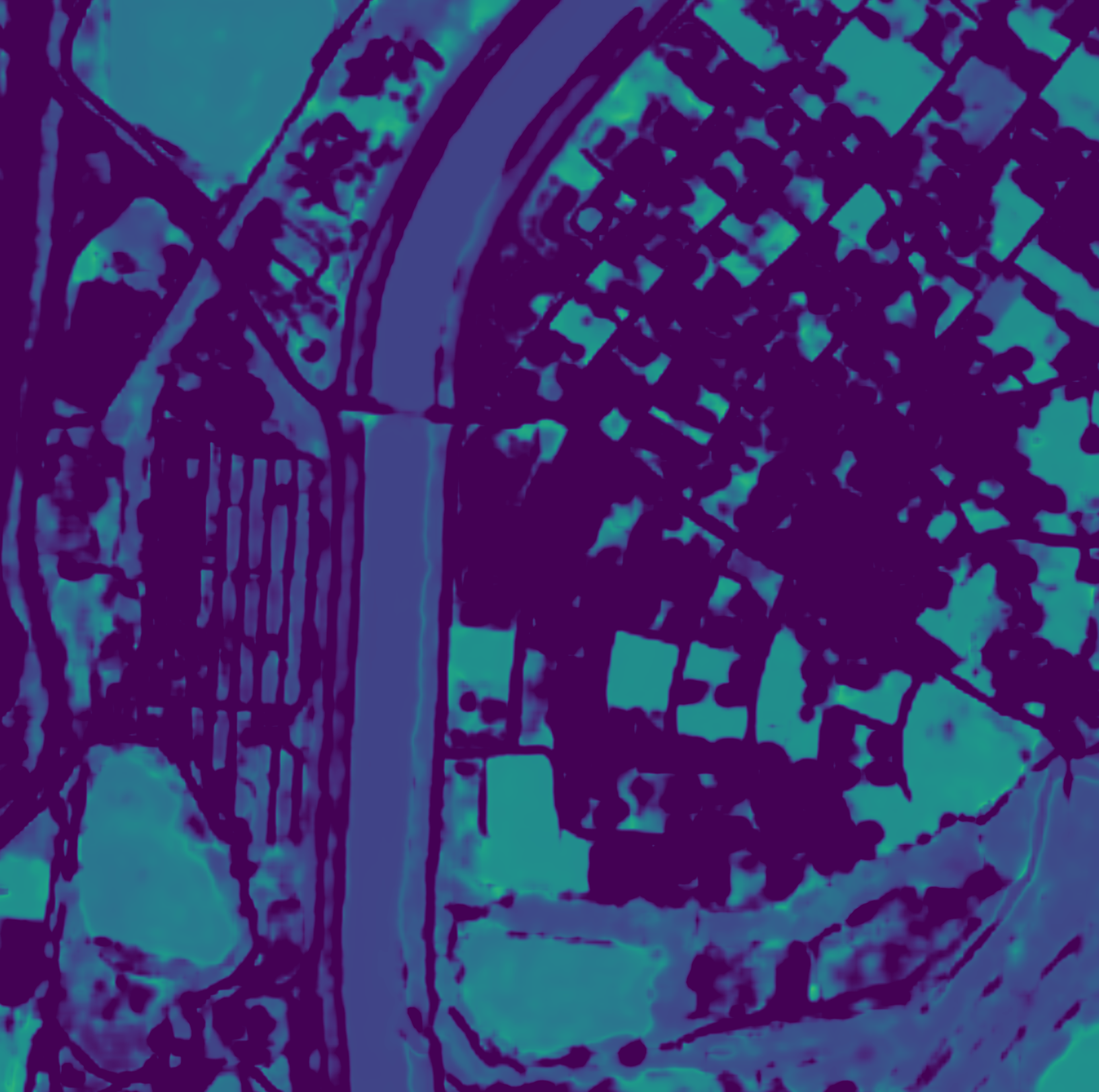} &
            \includegraphics[width=.15\linewidth]{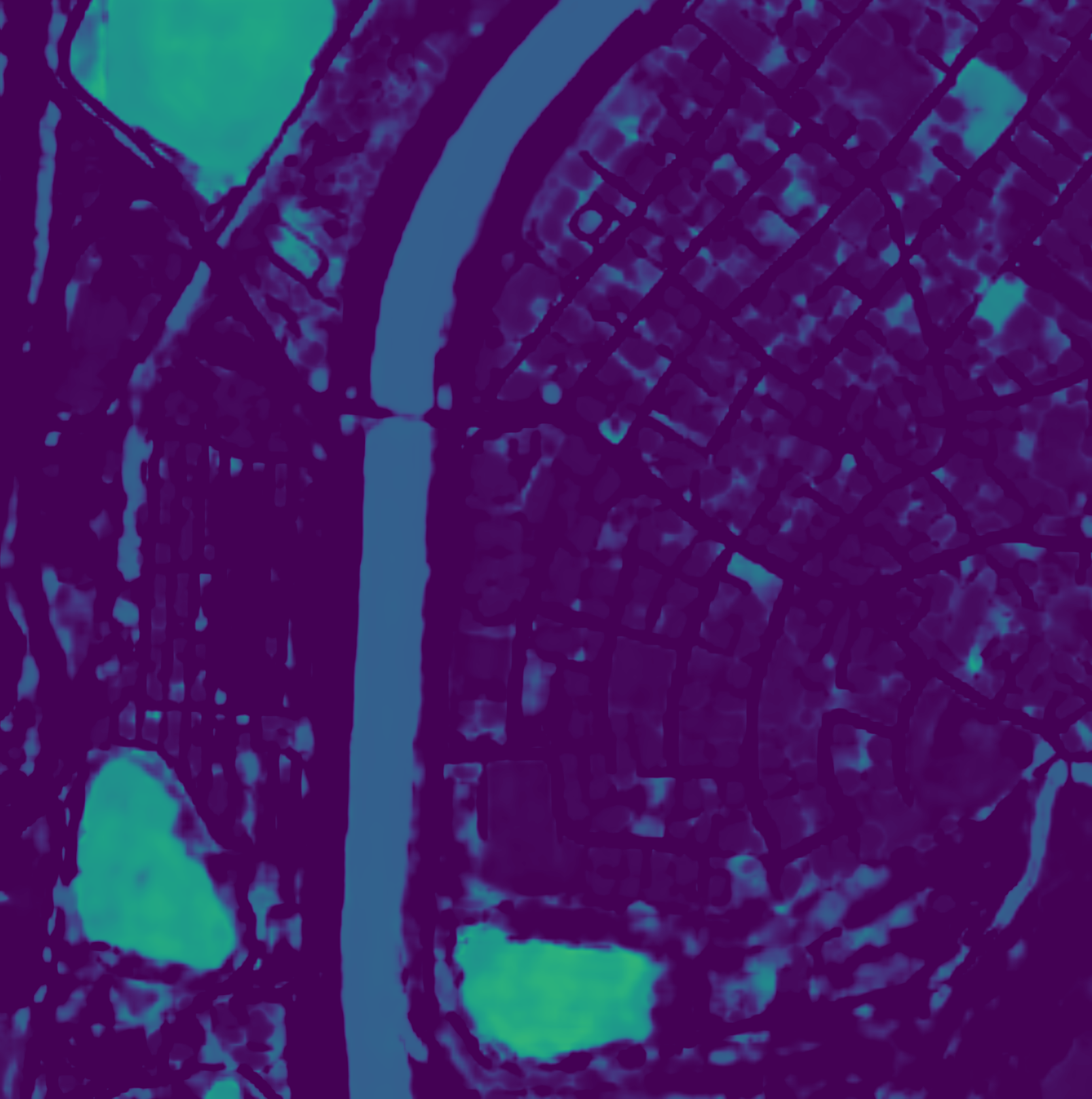} &
            \includegraphics[width=.15\linewidth]{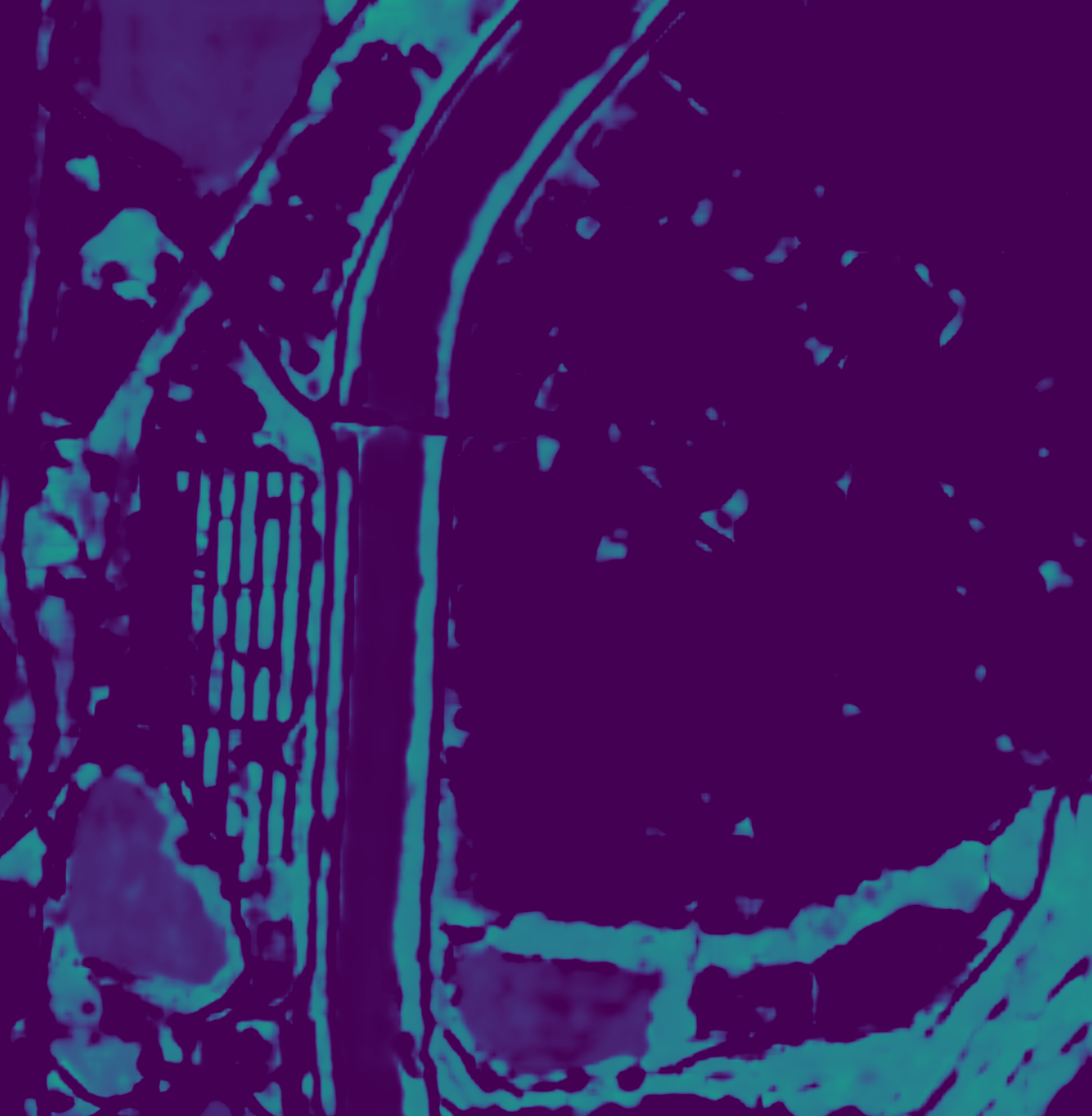}\\
                       \end{tabular}
%}
\end{minipage}
\caption{Image-text concepts alignment from~\cite{zermatten2023text}: given a model processing RGB images (a) and classifying landcover classes (b), the per pixel image features are aligned with a contrastive learning objective to the text concept vectors of GLoVE~\cite{glove}. Once aligned the image can be queried with text prompts for activation of concepts unseen during training, from semantically close to the landcover classes (c-d) to more far away, yet semantically related ones, such as `swimming' (e) or `squirrel'. Activations in (c-f) are normalized on a scale from 0 to 1, with dark blue colour for low values and light green for higher values.
%Labels for semantic segmentation maps : }\colorSquare{SkyBlue}~river, 
%\colorSquare{green}~agricultural areas, 
%\colorSquare{red}~buildings, 
%\colorSquare{OliveGreen}~forest, 
%\colorSquare{CornflowerBlue}~lake,  
%\colorSquare{lightgray}~roads ,
%\colorSquare{RoyalPurple}~vineyards. 
(\textcopyright, 2023 IEEE).
}  
\label{fig:val}
\end{figure*}

\label{sec:usercentric:lbi}
Besides label collection, interactivity can play a major role in opening access to EO information to virtually anyone. To reach such a goal, one must work on the access interface to the information itself. The engineering and modelling phase is a barrier insurmountable for many users, who cannot exploit the potential of EO because of technical impediments. Creating access in a more inclusive way, e.g. through natural language, would open EO to a much broader set of users, including decision-makers, the press and virtually all areas of society. Language opens EO access to society: methodologies ensuring this access via multimodal processing connecting natural language processing (NLP) models (understanding users' questions) and visual models (from computer vision, reviewed in the other sections) is critical. The interaction between language and remote sensing has become a very active area of research, whose main components are reviewed in~\cite{bashmal2023language} and are sketched in Fig.~\ref{fig:vlms}. In this section, we focus on two of them: visual question answering and image captioning / storytelling.
\begin{itemize}
\item 
Visual question answering~\cite{antol2015vqa} is a set of methodologies aiming at creating such links. The first approaches are showing proofs of concept for Sentinel resolution and VHR resolution imaging~\cite{lobry2020rsvqa} or emphasising changes in EO imagery \cite{yuan2022}. These first methods still have a large margin of progression, especially regarding the fusion strategies to mix language,   
the way to generate, learn and exploit from more realistic language,  
 the usage of large language models (LLM) from NLP (e.g. BERT, GPT)~\cite{wen2023visionlanguage} and the injection of EO context (or \emph{prompting}) in these models~\cite{Cha22pres}. These research lines will permit the development of digital assistants able to understand real questions from users and retrieve action-ready information from EO. The emergence of EO chatbots, scanning EO data and retrieving information on demand or dialoguing with users is an exciting avenue for both research and the industry. 

\item 
\label{sec:usercentric:story}
Besides answering user-specific questions, one could be interested in gathering general information about a scene for writing a journalistic story or simply having a summary of the image content for storytelling. Such image transcription in language form is often referred to as image summarisation and captioning. 
In this context, recent research has created large datasets, developed practical algorithms, and extended potential applications for RS image captioning tasks~\cite{sumbul2020sd,Kandala2022}.
Most existing remote sensing image captioning methods are based on encoder-decoder frameworks, which learn features from the input image, and then convert the encoded features into natural language descriptions. Attention mechanisms 
and transformers
 are extensively explored to this end. 
\end{itemize}
Beyond tasks where generating answers or a description of an image is the final goal, image/language alignment is also an interesting usage of language models, where language-based representations {(learned with vision language models such as CLIP~\cite{radford2021learning})} are used to build a bridge between semantic concepts across modalities: in \cite{mall2023remote}, representations from language models are aligned to those from remote sensing images by first aligning them to ground images co-located with the remote sensing ones. In \cite{zermatten2023text}, authors learn semantic segmentation models, where intermediate features are aligned to text embeddings from LLMs, allowing to activate image regions with respect to unseen concepts that are semantically close, but unseen during training, in a zero-shot manner {(Figure~\ref{fig:val} illustrates the out of vocabulary search abilities of a model initially meant for land cover mapping after its feature representations are aligned to those of a language model; since the features of water pixels now have representations close to those of the word `water', one can now search for new concepts by simply typing the corresponding words as text prompts).} In \cite{yuan2023rrsis}, authors introduce the task of referring remote sensing image segmentation (RRSIS), aiming at segmenting out the objects in a remote sensing image to which a given natural language expression refers. 

\subsection{Enhanced visualisation} 
The exploration of big multimodal data, including EO data multi-mission images, meta-data, or text calls for new visualisation techniques able to represent data in a way that eases interpretation and explanations, but also the discovery of unexpected information, anomalies and causal relations (see Section~\ref{sec:xai}). Succeeding in such exploration implies designing visualisation techniques, visual analytics, human-computer interaction, and augmented or virtual reality~\cite{sacha2017human}. 

Early geographic EO visualisation attempts were `virtual flights' on 3D terrains, rendered using 3D models from stereo or InSAR observations. Artefacts and noise removal, as well as multi-modal and multi-resolution representation in the RGB space, were the core issues. 
More recent visual data mining employs immersive visualisation~\cite{babaee2016immersive} of the EO information content with high user interaction. Immersive video environments like CAVE (a cube-sized room, where data are projected on 3 to 6 inner walls and the user is standing in its middle) or VR sets enriched with AI methods were used for multi-dimensional space adaptive projections. These visualisation methods and the analytics used interfaces must be predictive and adaptive, learning and anticipating the user behaviour and collaborating with them. This means understanding the user's intentions and context and establishing a dialogue to transform non-visual sensor data and information into understandable representations, potentially exploiting the approaches described in Section~\ref{sec:usercentric:lbi}.

\begin{tcolorbox}[width=\columnwidth,title={Promising directions for future research}]    
   \begin{itemize}
       \item[-] {Multimodal vision language models used to interactively query beyond concepts seen during training and adapt models across nomenclatures}
       \item[-] {Involving information beyond pixels in the reasoning process, for example involving knowledge graphs  mined with NLP tools~\cite{mi2022knowledge}}
       \item[-] {Interactive (chat)bots able to retrieve information on demand, following reasoning logic and feedback from users}
   \end{itemize}
   
\end{tcolorbox}  

%% file: sections/7_eosociety.tex
% !TEX root = ../main-GRSM.tex
\section{Earth observation and society: the growing relevance of ethics }
\label{sec:society}
 In this final section, we consider ethical issues linked to Artificial Intelligence, Earth Observation (AI4EO) and society under the following heads: 
(i) AI and Data Ethics
(ii) Ethical Opportunities for Sustainability: Environment and Health

\subsection{AI and Data Ethics} 
Ethical issues related to combined uses of AI and EO can emerge either in the context of  AI ethics or Data Ethics. The principles linked to each of these overlap. However, it is useful to understand them separately. 

\subsubsection{AI Ethics}  Jobin et al. \cite{RNY003} recently categorized AI ethics guidelines that most commonly appear in various national documents under 11 heads. Combining insights from Jobin et al. and four other relevant sources, Kochupillai et al.\cite{kochupillai2022earth} created 6 major categories of ethical duties (see Table~\ref{tab:AIethics}) and corresponding ethical consequences that are most relevant for the AI4EO community.

\begin{table*}[!t]
    \centering
    \small
      \caption{Six categories of ethical duties (from \cite{kochupillai2022earth}).}
    \begin{tabular}{|c|c | p{13cm}|}
    \hline
         1& Privacy & In the context of EO, this includes (i) Non-stigmatisation;
        (ii) respecting or protecting autonomy or freedom of persons and communities involved;
        (iii) Self-determination, including national sovereignty, and 
        (iv) ensuring proper data governance, including clear rules regarding data ownership and licensing.  \\
        \hline
         2& Honesty & Includes (i) transparency (including how representative the data is), (ii) explainability, and (iii) data veracity. It also requires good data segregation practices, e.g. preventing the overlap of test and training data.  \\
        \hline
         3& Integrity & Includes (i) ensuring technical robustness;
        (ii) proper uncertainty quantification and reporting of the limitations of the model; and
        (iii) safety/security, including ensuring national security (e.g. by selective disclosure of sensitive data) {as well as safety vulnerability of DL models to adversarial examples \cite{9442932} or patch attacks \cite{10119208}}. \\
        \hline
         4& Fairness & Includes. (i)Non-bias (including non-bias in training data);
        (ii) Non-discrimination and diversity (including adequately diverse and representative datasets)
        (iii) Socio-cultural sensitivities, e.g. while labelling data;
        (iv) Democratic creation of standards, e.g. when applying labels or using the data for policy decisions.\\
        \hline
         5& Responsibility & Includes (i) Human agency and oversight, or human-in-the-loop;
        (ii) duty of care, especially to understand corresponding ground realities;
        (iii) accountability; and
        (iv) ensuring social security and cohesion.\\
        \hline
         6& Sustainability & Includes (i) Scientific, social, and cultural sustainability (e.g. preventing ``colonialist science.''); and
        (ii) Environmental sustainability.\\
         \hline
    \end{tabular}
    \label{tab:AIethics}
\end{table*}

Several of these categories overlap with the principles of trustworthy AI compiled in the EU Ethics Guidelines for Trustworthy AI \cite{RNY004}. However, ethics guidelines, even when neatly categorized, may often appear vague or too high-level to EO scientists.\cite{hagendorff2020ethics} To gain a better and more practical understanding of ethics, it is useful to know that, at the simplest level, ethics is about minimising harm and maximising benefits using appropriate (fair, honest) means. This may not, however, be an easy or straightforward exercise in the context of everyday AI4EO research. In all instances involving ethical issues, several approaches can be utilized to reach an ethically acceptable decision or outcome. For example, dominant Western approaches to ethics consider the duties of the actor and consequences of the action/inaction (so-called deontological and consequentialist approaches to ethics) to determine their acceptability.\cite{thomas2015deontology}

Yet, these approaches may be impractical in emerging technologies such as AI4EO; scientists may not be able to clearly predict the future consequences of their research in the present and may not be fully aware of their duties. Recent work recommends combining approaches emerging from Eastern philosophy, particularly the intentions/desires of the individual researcher and/or the research project/program as a whole. Such considerations would include the evaluation of whether each action or research step corresponds with ethically well thought-out intentions or desires \cite{RNY005}.

 Consider the ethical principle of justice and fairness, which requires, for example, that appropriate measures be taken to ensure consistency and accuracy when applying sensitive data labels, such as slums. Even when accurate, such labels can lead to stigmatisation or even to uprooting people from their living spaces. Accordingly, taking into account the intention of the research or researcher, presumably, to improve living conditions in congested urban areas, ethics calls for more constructive labelling. This may also entail re-thinking research designs that can go beyond the labelling of slums and help uplift or improve the socio-economic conditions of marginalized segments. The principle of fairness, therefore, can be applied not only at the time of labelling data but even earlier in the research life-cycle, i.e. at the time of selecting research focus areas and research questions. Ethics is equally important at other stages of research - from allocating research funding to converting research findings to practical applications.  

Similarly, the issue of responsibility, which includes human agency or human-in-the-loop features in AI designs, needs careful consideration. While active involvement of human actors (diverse stakeholders) in every aspect of AI4EO research may not be possible, human agency and oversight must not be limited within the confines of the system and its narrow research question. Such oversight is also needed to understand the context in which the designed system is going to be implemented or introduced and the impact that it may have on real people, their lives, livelihoods, cultural sensitivities, and their environment. These insights are crucial to ensuring that the intended purpose of the research and development activity is actually going to be accomplished on the ground.

If the human agency and oversight requirement does not evolve considerably in the AI4EO research context,  intricate human realities, relationships, and expediencies can be overlooked or ignored, leading to significant ethically undesirable consequences.\cite{kochupillai2022earth} Ethics, therefore, calls for the humanisation of projects and supplementing any simulations and recommendations with multi-disciplinary social science and ground realities research. Such realities are not limited to ground truth or in situ data as understood in conventional EO language, but also include the socio-cultural, economic, and political context of the region that can make the consequences of implementing recommendations emerging from AI ethical or unethical. 

In this context, the rapid emergence of Large Language Models (LLMs) can increase the ethical risks related to inaccurate or outdated data, non-representative or biased datasets, errors, and misinformation,\cite{llmethics}, which can be further hidden by the size and scope of recent LLMs trained on large quantities of Web data. Moreover, AI4EO has a risk of leading to a kind of colonialist science, as Earth observation happens remotely. AI4EO approaches may give scientists who study places without ever going there a false impression that they know the place and its circumstances. This can lead to misunderstanding or a complete lack of understanding of local contexts and may even lead to disempowerment or discrimination against local communities, their rights over ancestral lands or occupations, their cultural values, and even their means of livelihood. Policy decisions that are made by relying on such science can also lead to incredible socio-economic inequity and violation of human rights.  

\subsubsection{Data Ethics}  Recently, the Data Ethics Commission set up by Germany’s Federal Government on 18 July 2018 \cite{dataeth2018de} highlighted 6 general ethical and legal principles from 8 guiding motifs. These general ethical principles are human dignity, self-determination, security, democracy, justice and solidarity, and sustainability. These principles signal the central relevance of the human-centred and value-oriented design of technology. However, finding a balance between potentially divergent principles is a complex matter requiring case-by-case attention. 

Consider the example of agriculture, where the EU Code of Conduct for agricultural data sharing emphasizes that “The farmer remains at the heart of the collection, processing, and management of agricultural data” \cite{RNY008}. The Code recognizes that farmers who share data have the right to know and participate in the uses to which their data is put. The challenge in EO and AI4EO research is that farmers' data is remotely collected, often without the knowledge of farmers. If the data collected is used merely to provide information to farmers, there is no ethical concern. However, when the data starts being used for the formulation of policies that (adversely) impact individual farmers' rights, the principles of the EU Code of Conduct will become very relevant. In fact, the use of AI in combination with EO in such instances will also trigger the application of the proposed EU AI Act, which may label such AI use cases as ``high risk'' and, therefore, subject to extensive regulation.

 \subsection{Ethical Opportunities for Sustainability: Environment and Health}
Beyond ethical issues, however, AI4EO research can also contribute to accomplishing various UN Sustainable Development Goals. This can range from assisting in the monitoring the compliance with or deviation from the SDGs or can actively support their accomplishment, e.g. to help expand or protect biodiversity. Recent research suggests that EO and Blockchain technology can be combined to support conservation efforts in Namibia \cite{RNY006}. EO can also help save lives during natural disasters \cite{RNY009}, help monitor climate change \cite{RNY011} and sea levels \cite{RNY012}.

Similarly, AI4EO applications also offer valuable benefits to public health. Epidemiological studies on the health effects of environmental hazards (such as air pollution and heatwaves) rely on observations from ground-based stations. The spatial distribution of the monitoring network is very heterogeneous and densely located in major cities, making it difficult to accurately represent the human exposure situation across the national territory. Recent AI4EO developments have focused on reconstructing historical air pollution levels at high spatial-temporal resolution using an extensive list of satellite-based data sources. Several single-learner \cite{rs12223803, atmos11030239, CHEN2019104934} and ensemble \cite{DI2019104909} ML models have also been explored to fill the gaps of air pollution concentrations, thus maximising the human exposure evidence. These spatiotemporal gap-filled environmental datasets are then linked to multiple health outcomes to estimate the associated health-risk exposure to environmental hazards and to quantify the total health burden (e.g., deaths) \cite{schneider2022,vicedo2021burden, URBAN2021111227}.

\begin{tcolorbox}[width=\columnwidth,title={Promising directions for future research}]    
   \begin{itemize}
       \item[-] {Research on means to conducting more inclusive and multi-disciplinary research in AI4EO, facilitating greater input of ground realities, including socio-cultural, economic and political realities}
       \item[-] {Research on means by which multi-cultural approaches can be used to design novel frameworks that facilitate interaction between EO scientists and ethics experts}
       \item[-] {Multi-disciplinary research for the creation of novel educational curricula that help EO scientists understand how to apply ethics in their day-to-day research}
   \end{itemize}
   
\end{tcolorbox}

%% file: main.bbl
\begin{thebibliography}{100}
\bibitem{RNY009}
P.~C. Oddo and J.~D. Bolton, ``The value of near real-time earth observations
  for improved flood disaster response,'' \emph{Frontiers in Environmental
  Science}, vol.~7, pp. 2--3, 2019.

\bibitem{RNY006}
D.~Oberhauser, ``Blockhain for environmental governance: Can smart contracts
  reinforce payments for ecosystem services in namibia,'' \emph{Frontiers in
  Blockchain}, vol.~2, 2019.

\bibitem{RNY011}
J.~Yang, P.~Gong, R.~Fu, M.~Zhang, J.~Chen, S.~Liang, B.~Xu, J.~Shi, and
  R.~Dickinson, ``The role of satellite remote sensing in climate change
  studies,'' \emph{Nature Climate Change}, vol.~3, no.~10, p. 875–883, 2013.

\bibitem{RNY012}
R.~Gens, ``Remote sensing of coastlines: Detection, extraction and
  monitoring,'' \emph{International Journal of Remote Sensing}, vol.~31, no.~7,
  p. 1819–1836, 2010.

\bibitem{zhu2017deep}
X.~X. Zhu, D.~Tuia, L.~Mou, G.-S. Xia, L.~Zhang, F.~Xu, and F.~Fraundorfer,
  ``Deep learning in remote sensing: A comprehensive review and list of
  resources,'' \emph{IEEE Geoscience and Remote Sensing Magazine}, vol.~5,
  no.~4, pp. 8--36, 2017.

\bibitem{persello22deep}
C.~Persello, J.~D. Wegner, R.~Haensch, D.~{Tuia}, P.~Ghamisi, M.~Koeva, and
  G.~Camps-Valls, ``Deep learning and earth observation to support the
  sustainable development goals: Current approaches, open challenges, and
  future opportunities,'' \emph{IEEE Geoscience and Remote Sensing Magazine},
  vol.~10, no.~2, pp. 172--200, 2022.

\bibitem{Cam20b}
G.~Camps-Valls, D.~Tuia, X.~X. Zhu, and M.~Reichstein, \emph{Deep learning for
  the Earth sciences: A comprehensive approach to remote sensing, climate
  science and geosciences}.\hskip 1em plus 0.5em minus 0.4em\relax Wiley and
  Sons, 2021.

\bibitem{Tui20grsm}
D.~Tuia, R.~Roscher, J.~D. Wegner, N.~Jacobs, X.~X. Zhu, and G.~Camps-Valls,
  ``Towards a collective agenda on ai for earth science data analysis,''
  \emph{IEEE Geoscience and Remote Sensing Magazine}, vol.~9, no.~2, pp.
  88--104, 2021.

\bibitem{schmitt2016data}
M.~Schmitt and X.~X. Zhu, ``Data fusion and remote sensing: An ever-growing
  relationship,'' \emph{IEEE Geoscience and Remote Sensing Magazine}, vol.~4,
  no.~4, pp. 6--23, 2016.

\bibitem{scheibenreif2022contrastive}
L.~Scheibenreif, M.~Mommert, and D.~Borth, ``Contrastive self-supervised data
  fusion for satellite imagery,'' \emph{ISPRS Annals of the Photogrammetry,
  Remote Sensing and Spatial Information Sciences}, vol.~3, pp. 705--711, 2022.

\bibitem{lefevre2017toward}
S.~Lef{\`e}vre, D.~Tuia, J.~D. Wegner, T.~Produit, and A.~S. Nassar, ``Toward
  seamless multiview scene analysis from satellite to street level,''
  \emph{Proceedings of the IEEE}, vol. 105, no.~10, pp. 1884--1899, 2017.

\bibitem{jongman2015early}
B.~Jongman, J.~Wagemaker, B.~R. Romero, and E.~C. De~Perez, ``Early flood
  detection for rapid humanitarian response: harnessing near real-time
  satellite and twitter signals,'' \emph{ISPRS International Journal of
  Geo-Information}, vol.~4, no.~4, pp. 2246--2266, 2015.

\bibitem{lang2023nee}
N.~Lang, W.~Jetz, K.~Schindler, and J.~D. Wegner, ``A high-resolution canopy
  height model of the earth,'' \emph{Nature Ecology \& Evolution}, 2023.

\bibitem{waldner2019needle}
F.~Waldner, Y.~Chen, R.~Lawes, and Z.~Hochman, ``Needle in a haystack: Mapping
  rare and infrequent crops using satellite imagery and data balancing
  methods,'' \emph{Remote Sensing of Environment}, vol. 233, p. 111375, 2019.

\bibitem{fei2006one}
L.~Fei-Fei, R.~Fergus, and P.~Perona, ``One-shot learning of object
  categories,'' \emph{IEEE Transactions on Pattern Analysis and Machine
  Intelligence}, vol.~28, no.~4, pp. 594--611, 2006.

\bibitem{hinton1999unsupervised}
G.~E. Hinton, T.~J. Sejnowski \emph{et~al.}, \emph{Unsupervised learning:
  foundations of neural computation}.\hskip 1em plus 0.5em minus 0.4em\relax
  MIT press, 1999.

\bibitem{brown2020language}
T.~Brown, B.~Mann, N.~Ryder, M.~Subbiah, J.~D. Kaplan, P.~Dhariwal,
  A.~Neelakantan, P.~Shyam, G.~Sastry, A.~Askell \emph{et~al.}, ``Language
  models are few-shot learners,'' \emph{Advances in Neural Information
  Processing Systems 33}, 2020.

\bibitem{chen2020simple}
T.~Chen, S.~Kornblith, M.~Norouzi, and G.~Hinton, ``A simple framework for
  contrastive learning of visual representations,'' in \emph{Proceedings of the
  37th International conference on machine learning, {ICML} 2020}.\hskip 1em
  plus 0.5em minus 0.4em\relax PMLR, 2020, pp. 1597--1607.

\bibitem{chen2021exploring}
X.~Chen and K.~He, ``Exploring simple siamese representation learning,'' in
  \emph{{IEEE} Conference on Computer Vision and Pattern Recognition, {CVPR}
  2021}, 2021, pp. 15\,750--15\,758.

\bibitem{he2022masked}
K.~He, X.~Chen, S.~Xie, Y.~Li, P.~Doll{\'a}r, and R.~Girshick, ``Masked
  autoencoders are scalable vision learners,'' in \emph{2022 IEEE/CVF
  Conference on Computer Vision and Pattern recognition}, 2022, pp.
  16\,000--16\,009.

\bibitem{tseng2023lightweight}
G.~Tseng, R.~Cartuyvels, I.~Zvonkov, M.~Purohit, D.~Rolnick, and H.~Kerner,
  ``Lightweight, pre-trained transformers for remote sensing timeseries,''
  \emph{arXiv preprint arXiv:2304.14065}, 2023.

\bibitem{irvin2023usat}
J.~Irvin, L.~Tao, J.~Zhou, Y.~Ma, L.~Nashold, B.~Liu, and A.~Y. Ng, ``Usat: A
  unified self-supervised encoder for multi-sensor satellite imagery,''
  \emph{arXiv preprint arXiv:2312.02199}, 2023.

\bibitem{sun2022ringmo}
X.~Sun, P.~Wang, W.~Lu, Z.~Zhu, X.~Lu, Q.~He, J.~Li, X.~Rong, Z.~Yang, H.~Chang
  \emph{et~al.}, ``Ringmo: A remote sensing foundation model with masked image
  modeling,'' \emph{IEEE Transactions on Geoscience and Remote Sensing},
  vol.~61, 2022.

\bibitem{ho2020denoising}
J.~Ho, A.~Jain, and P.~Abbeel, ``Denoising diffusion probabilistic models,'' in
  \emph{Advances in Neural Information Processing Systems}, vol.~33, 2020, pp.
  6840--6851.

\bibitem{rombach2022high}
R.~Rombach, A.~Blattmann, D.~Lorenz, P.~Esser, and B.~Ommer, ``High-resolution
  image synthesis with latent diffusion models,'' in \emph{IEEE/CVF Conference
  on Computer Vision and Pattern Recognition}, 2022, pp. 10\,684--10\,695.

\bibitem{bachmann2022multimae}
R.~Bachmann, D.~Mizrahi, A.~Atanov, and A.~Zamir, ``{MultiMAE}: Multi-modal
  multi-task masked autoencoders,'' in \emph{European Conference on Computer
  Vision}, 2022, pp. 348--367.

\bibitem{kirillov2023segment}
A.~Kirillov, E.~Mintun, N.~Ravi, H.~Mao, C.~Rolland, L.~Gustafson, T.~Xiao,
  S.~Whitehead, A.~C. Berg, W.-Y. Lo \emph{et~al.}, ``Segment anything,'' in
  \emph{IEEE/CVF International Conference on Computer Vision}, 2023, pp.
  4015--4026.

\bibitem{jakubik2023prithvi}
J.~Jakubik, S.~Roy, C.~E. Phillips, P.~Fraccaro, D.~Godwin, B.~Zadrozny,
  D.~Szwarcman, C.~Gomes, G.~Nyirjesy, B.~Edwards, D.~Kimura, N.~Simumba,
  L.~Chu, S.~K. Mukkavilli, D.~Lambhate, K.~Das, R.~Bangalore, D.~Oliveira,
  M.~Muszynski, K.~Ankur, M.~Ramasubramanian, I.~Gurung, S.~Khallaghi, H.~S.
  Li, M.~Cecil, M.~Ahmadi, F.~Kordi, H.~Alemohammad, M.~Maskey, R.~Ganti,
  K.~Weldemariam, and R.~Ramachandran, ``{Foundation Models for Generalist
  Geospatial Artificial Intelligence},'' \emph{arXiv preprint 2310.18660},
  2023.

\bibitem{satmae2022}
Y.~Cong, S.~Khanna, C.~Meng, P.~Liu, E.~Rozi, Y.~He, M.~Burke, D.~B. Lobell,
  and S.~Ermon, ``Satmae: Pre-training transformers for temporal and
  multi-spectral satellite imagery,'' in \emph{Advances in Neural Information
  Processing Systems 35: Annual Conference on Neural Information Processing
  Systems 2022, NeurIPS 2022}, 2022.

\bibitem{10.5555/3326943.3327027}
J.~Xu and Z.~Zhu, ``Reinforced continual learning,'' in \emph{Advances in
  Neural Information Processing Systems, NeurIPS}, 2018.

\bibitem{8898615}
O.~Tasar, Y.~Tarabalka, and P.~Alliez, ``Continual learning for dense labeling
  of satellite images,'' in \emph{International Geoscience and Remote Sensing
  Symposium, IGARSS}, 2019.

\bibitem{9184999}
N.~Ammour, Y.~Bazi, H.~Alhichri, and N.~Alajlan, ``Continual learning approach
  for remote sensing scene classification,'' \emph{IEEE Geoscience and Remote
  Sensing Letters}, vol.~19, pp. 1--5, 2022.

\bibitem{lenczner2022weaklysupervised}
G.~Lenczner, A.~Chan{-}Hon{-}Tong, N.~Luminari, and B.~{Le Saux},
  ``Weakly-supervised continual learning for class-incremental segmentation,''
  \emph{CoRR}, vol. abs/2201.01029, 2022.

\bibitem{Sri19}
S.~Srivastava, M.~Berman, M.~Blaschko, and D.~Tuia, ``Adaptive
  compression-based lifelong learning,'' in \emph{British Machine Vision
  Conference, BMVC}, 2019.

\bibitem{russwurm2022meteor}
M.~Russwurm, S.~Wang, B.~Kellenberger, R.~Roscher, and D.~{Tuia.},
  ``Meta-learning to address diverse earth observation problems across
  resolutions,'' \emph{Nature Communications Earth \& Environment}, vol.~5,
  no.~37, 2024.

\bibitem{chen2019closer}
W.-Y. Chen, Y.-C. Liu, Z.~Kira, Y.-C.~F. Wang, and J.-B. Huang, ``A closer look
  at few-shot classification,'' in \emph{International Conference on Learning
  Representations, ICLR}, 2019.

\bibitem{sun2016deep}
B.~Sun and K.~Saenko, ``Deep coral: Correlation alignment for deep domain
  adaptation,'' in \emph{European Conference on Computer Vision, ECCV
  Workshops}, 2016.

\bibitem{xu2022the}
M.~Xu, M.~Wu, K.~Chen, C.~Zhang, and J.~Guo, ``The eyes of the gods: {A} survey
  of unsupervised domain adaptation methods based on remote sensing data,''
  \emph{Remote Sensing}, vol.~14, no.~17, p. 4380, 2022.

\bibitem{sung2018learning}
F.~Sung, Y.~Yang, L.~Zhang, T.~Xiang, P.~H. Torr, and T.~M. Hospedales,
  ``Learning to compare: Relation network for few-shot learning,'' in
  \emph{Computer Vision and Pattern Recognition, CVPR}, 2018.

\bibitem{xian2017zero}
Y.~Xian, B.~Schiele, and Z.~Akata, ``Zero-shot learning -- the good, the bad
  and the ugly,'' in \emph{Computer Vision and Pattern Recognition, CVPR},
  2017.

\bibitem{brazdil2022metalearning}
P.~Brazdil, J.~N. van Rijn, C.~Soares, and J.~Vanschoren, \emph{Metalearning:
  Applications to Automated Machine Learning and Data Mining}, 2nd~ed.\hskip
  1em plus 0.5em minus 0.4em\relax Springer, 2022.

\bibitem{huisman2021survey}
M.~Huisman, J.~N. van Rijn, and A.~Plaat, ``A survey of deep meta-learning,''
  \emph{Artificial Intelligence Review}, vol.~54, no.~6, pp. 4483--4541, 2021.

\bibitem{white2023neural}
C.~White, M.~Safari, R.~Sukthanker, B.~Ru, T.~Elsken, A.~Zela, D.~Dey, and
  F.~Hutter, ``Neural architecture search: Insights from 1000 papers,''
  \emph{arXiv preprint arXiv:2301.08727}, 2023.

\bibitem{hutter2019automated}
F.~Hutter, L.~Kotthoff, and J.~Vanschoren, \emph{Automated machine learning:
  methods, systems, challenges}.\hskip 1em plus 0.5em minus 0.4em\relax
  Springer, 2019.

\bibitem{ying2019bench}
C.~Ying, A.~Klein, E.~Christiansen, E.~Real, K.~Murphy, and F.~Hutter,
  ``Nas-bench-101: Towards reproducible neural architecture search,'' in
  \emph{Proceedings of the 36th International Conference on Machine
  Learning,{ICML} 2019}, ser. Proceedings of Machine Learning Research,
  vol.~97, 2019, pp. 7105--7114.

\bibitem{sumbul2019bigearthnet}
G.~Sumbul, M.~Charfuelan, B.~Demir, and V.~Markl, ``Bigearthnet: A large-scale
  benchmark archive for remote sensing image understanding,'' in
  \emph{International Geoscience and Remote Sensing Symposium, {IGARSS}}, 2019.

\bibitem{Hutter2014}
F.~Hutter, H.~H. Hoos, and K.~Leyton-Brown, ``An efficient approach for
  assessing hyperparameter importance,'' in \emph{Proceedings of the 31th
  International Conference on Machine Learning, {ICML} 2014}, 2014, pp.
  754--762.

\bibitem{perrone2019learning}
V.~Perrone, H.~Shen, M.~W. Seeger, C.~Archambeau, and R.~Jenatton, ``Learning
  search spaces for bayesian optimization: Another view of hyperparameter
  transfer learning,'' in \emph{Advances in Neural Information Processing
  Systems 32}, 2019, pp. 12\,751--12\,761.

\bibitem{Rijn2018}
J.~N. van Rijn and F.~Hutter, ``Hyperparameter importance across datasets,'' in
  \emph{Proceedings of the 24th ACM SIGKDD International Conference on
  Knowledge Discovery \& Data Mining, KDD 2018}.\hskip 1em plus 0.5em minus
  0.4em\relax ACM, 2018, pp. 2367--2376.

\bibitem{mohr2022learning}
F.~Mohr and J.~N. van Rijn, ``Learning curves for decision making in supervised
  machine learning -- a survey,'' \emph{CoRR}, vol. abs/2201.12150, 2022.

\bibitem{mohr2023fast}
------, ``Fast and informative model selection using learning curve
  cross-validation,'' \emph{IEEE Transactions on Pattern Analysis and Machine
  Intelligence}, vol.~45, no.~8, pp. 9669--9680, 2023.

\bibitem{konig2024accelerating}
M.~König, H.~H. Hoos, and J.~N. van Rijn, ``Accelerating adversarially robust
  model selection for deep neural networks via racing,'' in \emph{Proceedings
  of the 38th {AAAI} Conference on Artificial Intelligence ({AAAI}-24)}.\hskip
  1em plus 0.5em minus 0.4em\relax {AAAI} Press, 2024, pp. 21\,267--21\,275.

\bibitem{silva2020opportunities}
S.~H. Silva and P.~Najafirad, ``Opportunities and challenges in deep learning
  adversarial robustness: A survey,'' \emph{arXiv preprint arXiv:2007.00753},
  2020.

\bibitem{palacios2021automated}
N.~R. Palacios~Salinas, M.~Baratchi, J.~N. van Rijn, and A.~Vollrath,
  ``Automated machine learning for satellite data: integrating remote sensing
  pre-trained models into automl systems,'' in \emph{Machine Learning and
  Knowledge Discovery in Databases}, ser. Lecture Notes in Computer Science,
  vol. 12979.\hskip 1em plus 0.5em minus 0.4em\relax Springer, 2021, pp.
  447--462.

\bibitem{wasala2024autosr4eo}
J.~Wasala, S.~Marselis, L.~Arp, H.~Hoos, N.~Long\'{e}p\'{e}, and M.~Baratchi,
  ``Autosr4eo: An automl approach to super-resolution for earth observation
  images,'' \emph{Remote Sensing}, vol.~16, no.~3, 2024.

\bibitem{gavsparovic2018fusion}
M.~Ga{\v{s}}parovi{\'c}, D.~Medak, I.~Pila{\v{s}}, L.~Jurjevi{\'c}, and
  I.~Balenovi{\'c}, ``Fusion of {Sentinel-2} and {Planetscope} imagery for
  vegetation detection and monitoring,'' \emph{International Archives of the
  Photogrammetry, Remote Sensing \& Spatial Information Sciences}, vol.~42,
  no.~1, pp. 155--160, 2018.

\bibitem{chollet2019}
F.~Chollet, ``On the measure of intelligence,'' \emph{arXiv preprint
  arXiv:1911.01547}, 2019.

\bibitem{sudmanns_big_2020}
M.~Sudmanns, D.~Tiede, S.~Lang, H.~Bergstedt, G.~Trost, H.~Augustin,
  A.~Baraldi, and T.~Blaschke, ``Big earth data: disruptive changes in earth
  observation data management and analysis?'' \emph{International Journal of
  Digital Earth}, vol.~13, no.~7, pp. 832--850, 2020.

\bibitem{Wall2021}
A.~Wall, B.~Deiseroth, E.~Tzirita~Zacharatou, J.-A. Quian\'{e}-Ruiz, B.~Demir,
  and V.~Markl, ``Agora-eo: A unified ecosystem for earth observation - a
  vision for boosting eo data literacy,'' \emph{Big Data from Space
  Conference}, 2021.

\bibitem{Traub2020}
J.~Traub, Z.~Kaoudi, J.-A. Quian\'{e}-Ruiz, and V.~Markl, ``Agora: Bringing
  together datasets, algorithms, models and more in a unified ecosystem
  [vision],'' \emph{SIGMOD Record}, vol.~49, no.~4, p. 6–11, 2021.

\bibitem{federatedlearning}
%\BIBentryALTinterwordspacing
J.~Kone\v{c}n\'{y}, H.~B. McMahan, F.~X. Yu, P.~Richtarik, A.~T. Suresh, and
  D.~Bacon, ``Federated learning:strategies for improving communication
  efficiency,'' in \emph{Neural Information Processing Systems, NeurIPS
  workshops}, 2016. [Online]. Available: \url{https://arxiv.org/abs/1610.05492}
%\BIBentrySTDinterwordspacing

\bibitem{federatedlearningblog}
``Federated learning: Collaborative machine learning without centralized
  training data,'' Google AI Blog:
  \url{https://ai.googleblog.com/2017/04/federated-learning-collaborative.html},
  accessed: 2022-02-28.

\bibitem{federatedlearning2017}
B.~McMahan, E.~Moore, D.~Ramage, S.~Hampson, and B.~A. y~Arcas,
  ``Communication-efficient learning of deep nnetworks from decentralized
  data,'' in \emph{International Conference on Artificial Intelligence and
  Statistics, {AISTATS}}, vol.~54, 2017, pp. 1273--1282.

\bibitem{federatedlearningNon-IID}
Y.~Zhao, M.~Li, L.~Lai, N.~Suda, D.~Civin, and V.~Chandra, ``Federated learning
  with non-iid data,'' \emph{arXiv preprint arXiv:1806.00582}, 2018.

\bibitem{10282873}
B.~Büyüktaş, G.~Sumbul, and B.~Demir, ``Learning across decentralized
  multi-modal remote sensing archives with federated learning,'' in
  \emph{{IEEE} International Geoscience and Remote Sensing Symposium, {IGARSS}
  2023}.\hskip 1em plus 0.5em minus 0.4em\relax {IEEE}, 2023, pp. 4966--4969.

\bibitem{activeloop}
``Hub: A common database for ai,'' \url{https://www.activeloop.ai/}, accessed:
  2022-02-28.

\bibitem{onnx}
``Open neural network exchange,'' \url{https://onnx.ai/}, accessed: 2022-02-28.

\bibitem{Vanschorenetal2014}
J.~Vanschoren, J.~N. van Rijn, B.~Bischl, and L.~Torgo, ``Openml: networked
  science in machine learning,'' \emph{SIGKDD Explorations}, vol.~15, pp.
  49--60, 2014.

\bibitem{Ziajaetal}
M.~Ziaja, P.~Bosowski, M.~Myller, G.~Gajoch, M.~Gumiela, J.~Protich, K.~Borda,
  D.~Jayaraman, R.~Dividino, and J.~Nalepa, ``Benchmarking deep learning for
  on-board space applications,'' \emph{Remote Sensing}, vol.~13, no.~19, 2021.

\bibitem{gernigon2023low}
C.~Gernigon, S.-I. Filip, O.~Sentieys, C.~Coggiola, and M.~Bruno,
  ``Low-precision floating-point for efficient on-board deep neural network
  processing,'' in \emph{2023 European Data Handling \& Data Processing
  Conference (EDHPC)}.\hskip 1em plus 0.5em minus 0.4em\relax IEEE, 2023, pp.
  1--8.

\bibitem{murphy2021machine}
J.~Murphy, J.~E. Ward, and B.~M. Namee, ``Machine learning in space: {A} review
  of machine learning algorithms and hardware for space applications,'' in
  \emph{The 29th Irish Conference on Artificial Intelligence and Cognitive
  Science 2021, Dublin, Republic of Ireland, December 9-10, 2021}, ser. {CEUR}
  Workshop Proceedings, vol. 3105.\hskip 1em plus 0.5em minus 0.4em\relax
  CEUR-WS.org, 2021, pp. 72--83.

\bibitem{guerrisi2023artificial}
G.~Guerrisi, F.~D. Frate, and G.~Schiavon, ``Artificial intelligence based
  on-board image compression for the {\(\Phi\)}-sat-2 mission,'' \emph{{IEEE}
  Journal of Selected Topics in Applied Earth Observations and Remote Sensing},
  vol.~16, pp. 8063--8075, 2023.

\bibitem{li2021dynamic}
C.~Li, G.~Wang, B.~Wang, X.~Liang, Z.~Li, and X.~Chang, ``Dynamic slimmable
  network,'' in \emph{2021 IEEE/CVF Conference on Computer Vision and Pattern
  Recognition}, 2021, pp. 8607--8617.

\bibitem{page2023developing}
C.~Page, K.~Cahoy, and E.~Gizzi, ``Developing intelligent space systems: A
  survey \& rubric for future missions,'' in \emph{37th Annual Small Satellite
  Conference}, 2023.

\bibitem{arvor2019ontologies}
D.~Arvor, M.~Belgiu, Z.~Falomir, I.~Mougenot, and L.~Durieux, ``Ontologies to
  interpret remote sensing images: why do we need them?'' \emph{GIScience \&
  remote sensing}, vol.~56, no.~6, pp. 911--939, 2019.

\bibitem{ras2022explainable}
G.~Ras, N.~Xie, M.~Van~Gerven, and D.~Doran, ``Explainable deep learning: A
  field guide for the uninitiated,'' \emph{Journal of Artificial Intelligence
  Research}, vol.~73, pp. 329--396, 2022.

\bibitem{rudin2019stop}
C.~Rudin, ``Stop explaining black box machine learning models for high stakes
  decisions and use interpretable models instead,'' \emph{Nature machine
  intelligence}, vol.~1, no.~5, pp. 206--215, 2019.

\bibitem{Pearl2000}
J.~Pearl, \emph{Causality: Models, Reasoning and Inference}, 2nd~ed.\hskip 1em
  plus 0.5em minus 0.4em\relax Cambridge University Press, 2009.

\bibitem{camps2018physics}
G.~Camps-Valls, L.~Martino, D.~H. Svendsen, M.~Campos-Taberner,
  J.~Mu{\~n}oz-Mar{\'\i}, V.~Laparra, D.~Luengo, and F.~J. Garc{\'\i}a-Haro,
  ``Physics-aware gaussian processes in remote sensing,'' \emph{Applied Soft
  Computing}, vol.~68, pp. 69--82, 2018.

\bibitem{Reichstein19nat}
M.~Reichstein, G.~Camps-Valls, B.~Stevens, M.~Jung, J.~Denzler, N.~Carvalhais,
  and Prabhat, ``Deep learning and process understanding for data-driven earth
  system science,'' \emph{Nature}, vol. 566, pp. 195--204, 2019.

\bibitem{CampsValls23physcausaldiscovery}
G.~Camps-Valls, A.~Gerhardus, U.~Ninad, G.~Varando, G.~Martius,
  E.~Balaguer-Ballester, R.~Vinuesa, E.~Diaz, L.~Zanna, and J.~Runge,
  ``Discovering causal relations and equations from data,'' \emph{Physics
  Reports}, vol. 1044, pp. 1--68, 2023.

\bibitem{Jackson1999}
P.~Jackson, \emph{Introduction to Expert Systems}, 3rd~ed.\hskip 1em plus 0.5em
  minus 0.4em\relax Addison-Wesley, 1999.

\bibitem{Bohanec2022}
M.~Bohanec, ``Dex (decision expert): A qualitative hierarchical multi-criteria
  method,'' in \emph{Multiple Criteria Decision Making}.\hskip 1em plus 0.5em
  minus 0.4em\relax Springer Singapore, 2022, pp. 39--78.

\bibitem{Debeljaketal2019}
M.~Debeljak, A.~Trajanov, V.~Kuzmanovski, J.~Schr\"{o}der, T.~Sand{\'{e}}n,
  H.~Spiegel, D.~P. Wall, M.~V. de~Broek, M.~Rutgers, F.~Bampa, R.~E. Creamer,
  and C.~B. Henriksen, ``A field-scale decision support system for assessment
  and management of soil functions,'' \emph{Frontiers in Environmental
  Science}, vol.~7, p. 115, 2019.

\bibitem{OECD}
``Earth observation for decision making,''
  \url{http://www.oecd.org/environment/indicators-modelling-outlooks/earth-observation-for-decision-making.htm},
  accessed: 2022-03-22.

\bibitem{Vinuesaetal2020}
R.~Vinuesa, H.~Azizpour, I.~Leite, M.~Balaam, V.~Dignum, S.~Domisch,
  A.~Fell\"{a}nder, S.~D. Langhans, M.~Tegmark, and F.~F. Nerini, ``The role of
  artificial intelligence in achieving the sustainable development goals,''
  \emph{Nature Communications}, vol.~11, no.~1, pp. 1--10, 2020.

\bibitem{Mitchelletall1986}
T.~M. Mitchell, R.~M. Keller, and S.~T. Kedar-Cabelli, ``Explanation-based
  generalization: A unifying view,'' \emph{Machine Learning}, vol.~1, no.~1,
  pp. 47--80, 1986.

\bibitem{LavracandDzeroski1994}
N.~{Lavra\v{c}} and S.~{D\v{z}eroski}, \emph{Inductive logic programming.
  Techniques and applications}, ser. Ellis Horwood series in artificial
  intelligence.\hskip 1em plus 0.5em minus 0.4em\relax Ellis Horwood, 1994.

\bibitem{DzeroskiandLavrac2001}
S.~D{\v{z}}eroski and N.~Lavra{\v{c}},
  \emph{Relational data mining}.\hskip 1em plus
  0.5em minus 0.4em\relax Springer, 2001.

\bibitem{GetoorandTaskar2007}
L.~Getoor and B.~Taskar, \emph{Introduction to Statistical Relational
  Learning}.\hskip 1em plus 0.5em minus 0.4em\relax The MIT Press, 2007.

\bibitem{DeReadtetal2016}
L.~D. Raedt, K.~Kersting, S.~Natarajan, and D.~Poole, ``Statistical relational
  artificial intelligence: Logic, probability, and computation,''
  \emph{Synthesis Lectures on Artificial Intelligence and Machine Learning},
  vol.~10, no.~2, pp. 1--189, 2016.

\bibitem{SillaandFreitas2010}
C.~N. Silla and A.~A. Freitas, ``A survey of hierarchical classification across
  different application domains,'' \emph{Data Mining and Knowledge Discovery},
  vol.~22, no. 1-2, pp. 31--72, 2010.

\bibitem{Yangetal2017}
H.~Yang, S.~Li, J.~Chen, X.~Zhang, and S.~Xu, ``The standardization and
  harmonization of land cover classification systems towards harmonized
  datasets: A review,'' \emph{ISPRS International Journal of Geo-Information},
  vol.~6, no.~5, p. 154, 2017.

\bibitem{Stivaktakisetal2019}
R.~Stivaktakis, G.~Tsagkatakis, and P.~Tsakalides, ``Deep learning for
  multilabel land cover scene categorization using data augmentation,''
  \emph{IEEE Geoscience and Remote Sensing Letters}, vol.~16, no.~7, pp.
  1031--1035, 2019.

\bibitem{sumbul2020deep}
G.~Sumbul and B.~Dem{\.I}r, ``A deep multi-attention driven approach for
  multi-label remote sensing image classification,'' \emph{IEEE Access},
  vol.~8, pp. 95\,934--95\,946, 2020.

\bibitem{mollenbrok2023deep}
L.~M{\"o}llenbrok, G.~Sumbul, and B.~Demir, ``Deep active learning for
  multi-label classification of remote sensing images,'' \emph{IEEE Geoscience
  and Remote Sensing Letters}, 2023.

\bibitem{Tui10b}
D.~{Tuia}, J.~Mu{\~n}oz-Mar{\'\i}, M.~Kanevski, and G.~Camps-Valls,
  ``Structured output {SVM} for remote sensing image classification,''
  \emph{Journal of Signal Processing Systems}, vol.~65, no.~3, pp. 457--468,
  2011.

\bibitem{turkoglu2021crop}
M.~O. Turkoglu, S.~D'Aronco, G.~Perich, F.~Liebisch, C.~Streit, K.~Schindler,
  and J.~D. Wegner, ``Crop mapping from image time series: deep learning with
  multi-scale label hierarchies,'' \emph{Remote Sensing of Environment}, vol.
  264, p. 112603, 2021.

\bibitem{Dzeroskietal2010}
S.~D{\v{z}}eroski, B.~Goethals, and P.~Panov, Eds., \emph{Inductive Databases
  and Constraint-Based Data Mining}.\hskip 1em plus 0.5em minus 0.4em\relax
  Springer New York, 2010.

\bibitem{DzeroskiandTodorovski2007}
S.~D{\v{z}}eroski and L.~Todorovski, Eds., \emph{Computational Discovery of
  Scientific Knowledge}.\hskip 1em plus 0.5em minus 0.4em\relax Springer Berlin
  Heidelberg, 2007.

\bibitem{Simidjievskietal2020}
N.~Simidjievski, L.~Todorovski, J.~Kocijan, and S.~Dzeroski, ``Equation
  discovery for nonlinear system identification,'' \emph{{IEEE} Access},
  vol.~8, pp. 29\,930--29\,943, 2020.

\bibitem{Wilkinsonetal2016}
M.~D. Wilkinson, ``The fair guiding principles for scientific data management
  and stewardship,'' \emph{Scientific Data}, vol.~3, p. 160018, 2016.

\bibitem{Giulianietal2021}
G.~Giuliani, H.~Cazeaux, P.-Y. Burgi, C.~Poussin, J.-P. Richard, and
  B.~Chatenoux, ``Swissenveo: A fair national environmental data repository for
  earth observation open science,'' \emph{Data Science Journal}, vol.~20, 2021.

\bibitem{grujdin2023self}
I.~Grujdin and M.~Datcu, ``Self-learning ontology for natural hazards,'' in
  \emph{IGARSS 2023-2023 IEEE International Geoscience and Remote Sensing
  Symposium}, 2023, pp. 2227--2230.

\bibitem{Runge19natcom}
J.~Runge, S.~Bathiany, E.~Bollt, G.~Camps-Valls, D.~Coumou, E.~Deyle,
  C.~Glymour, M.~Kretschmer, M.~Mahecha, J.~Mu{\~n}oz-Mar\'i, E.~van Nes,
  J.~Peters, R.~Quax, M.~Reichstein, M.~Scheffer, B.~Sch\"olkopf, P.~Spirtes,
  G.~Sugihara, J.~Sun, K.~Zhang, and J.~Zscheischler, ``Inferring causation
  from time series with perspectives in earth system sciences,'' \emph{Nature
  Communications}, vol.~10, no. 2553, 2019.

\bibitem{Runge23causalreview}
J.~Runge, A.~Gerhardus, G.~Varando, V.~Eyring, and G.~Camps-Valls, ``Causal
  inference for time series,'' \emph{Nature Reviews Earth \& Environment},
  vol.~10, p. 2553, 2023.

\bibitem{SamArXiv20}
W.~Samek, G.~Montavon, S.~Lapuschkin, C.~J. Anders, and K.-R. M{\'u}ller,
  ``Toward interpretable machine learning: Transparent deep neural networks and
  beyond,'' \emph{arXiv preprint arXiv:2003.07631}, 2020.

\bibitem{roscher2020explainable}
R.~Roscher, B.~Bohn, M.~F. Duarte, and J.~Garcke, ``Explainable machine
  learning for scientific insights and discoveries,'' \emph{IEEE Access},
  vol.~8, pp. 42\,200--42\,216, 2020.

\bibitem{Ola19xai}
A.~Wolanin, G.~Mateo-Garcia, G.~Camps-Valls, L.~Gomez-Chova, M.~Meroni,
  G.~Duveiller, Y.~Liangzhi, and L.~Guanter, ``Estimating and understanding
  crop yields with explainable deep learning in the indian wheat belt,''
  \emph{Environmental Research Letters}, vol.~15, no.~2, pp. 1--12, 2020.

\bibitem{Johnson20sakame}
J.~E. Johnson, V.~Laparra, A.~P\'{e}rez-Suay, M.~Mahecha, and G.~Camps-Valls,
  ``Kernel methods and their derivatives: Concept and perspectives for the
  earth system sciences,'' \emph{PLOS One}, vol.~15, no.~10, p. p.e0235885,
  2020.

\bibitem{Marcos2019}
D.~Marcos, S.~Lobry, and D.~Tuia, ``Semantically interpretable activation maps:
  What-where-how explanations within cnns,'' in \emph{International Conference
  on Computer Vision Workshops, {ICCV} Workshops}, 2019, pp. 4207--4215.

\bibitem{russwurm2020self}
M.~Ru{\ss}wurm and M.~K{\"o}rner, ``Self-attention for raw optical satellite
  time series classification,'' \emph{ISPRS Journal of Photogrammetry and
  Remote Sensing}, vol. 169, pp. 421--435, 2020.

\bibitem{levering2021relation}
A.~Levering, D.~Marcos, and D.~Tuia, ``On the relation between landscape beauty
  and land cover: A case study in the uk at sentinel-2 resolution with
  interpretable ai,'' \emph{ISPRS journal of Photogrammetry and Remote
  Sensing}, vol. 177, pp. 194--203, 2021.

\bibitem{nguyen2022mapping}
T.-A. Nguyen, B.~Kellenberger, and D.~Tuia, ``Mapping forest in the swiss alps
  treeline ecotone with explainable deep learning,'' \emph{Remote Sensing of
  Environment}, vol. 281, p. 113217, 2022.

\bibitem{stomberg2021jungle}
T.~Stomberg, I.~Weber, M.~Schmitt, and R.~Roscher, ``Jungle-net: Using
  explainable machine learning to gain new insights into the appearance of
  wilderness in satellite imagery,'' \emph{ISPRS Annals of the Photogrammetry,
  Remote Sensing and Spatial Information Sciences}, vol.~3, pp. 317--324, 2021.

\bibitem{Diaz21rccm}
E.~Diaz, J.~E. Adsuara, A.~Moreno-Martinez, M.~Piles, and G.~Camps-Valls,
  ``Inferring causal relations from observational long-term carbon and water
  fluxes records,'' \emph{Sientific Reports}, vol.~12, p. 1610, 2022.

\bibitem{Peters18}
J.~Peters, D.~Janzing, and B.~Sch{\"o}lkopf, \emph{Elements of Causal Inference
  - Foundations and Learning Algorithms}, ser. Adaptive Computation and Machine
  Learning Series.\hskip 1em plus 0.5em minus 0.4em\relax MIT Press, 2017.

\bibitem{varandolearning}
G.~Varando, M.-A. Fern\'{a}ndez-Torres, and G.~Camps-Valls, ``Learning granger
  causal feature representations,'' in \emph{International Conference on
  Machine Learning, ICML Workshops}, 2021.

\bibitem{svendsen17jgp}
D.~H. Svendsen, L.~Martino, M.~Campos-Taberner, F.~J. Garc{\'\i}a-Haro, and
  G.~Camps-Valls, ``Joint gaussian processes for biophysical parameter
  retrieval,'' \emph{IEEE Transactions on Geoscience and Remote Sensing},
  vol.~56, no.~3, pp. 1718--1727, 2017.

\bibitem{Samek2019}
W.~Samek, G.~Montavon, A.~Vedaldi, L.~K. Hansen, and K.-R. M{\"u}ller,
  \emph{Explainable AI: Interpreting, Explaining and Visualizing Deep
  Learning}, ser. Lecture Notes in Computer Science.\hskip 1em plus 0.5em minus
  0.4em\relax Springer, 2019, vol. 11700.

\bibitem{von2019informed}
L.~Von~Rueden, S.~Mayer, J.~Garcke, C.~Bauckhage, and J.~Schuecker, ``Informed
  machine learning--towards a taxonomy of explicit integration of knowledge
  into machine learning,'' \emph{Learning}, vol.~18, pp. 19--20, 2019.

\bibitem{camps2016survey}
G.~Camps-Valls, J.~Verrelst, J.~Mu{\~n}oz-Mar\'i, V.~Laparra,
  F.~Mateo-Jim\'enez, and J.~Gomez-Dans, ``A survey on gaussian processes for
  earth-observation data analysis: A comprehensive investigation,'' \emph{IEEE
  Geoscience and Remote Sensing Magazine}, vol.~4, no.~2, pp. 58--78, 2016.

\bibitem{camps2019perspective}
G.~Camps-Valls, D.~Sejdinovic, J.~Runge, and M.~Reichstein, ``A perspective on
  gaussian processes for earth observation,'' \emph{National Science Review},
  vol.~6, no.~4, pp. 616--618, 2019.

\bibitem{del2004neural}
F.~Del~Frate and D.~Solimini, ``On neural network algorithms for retrieving
  forest biomass from {SAR} data,'' \emph{IEEE Transactions on Geoscience and
  Remote Sensing}, vol.~42, no.~1, pp. 24--34, 2004.

\bibitem{sellitto}
P.~Sellitto, F.~D. Frate, D.~Solimini, and S.~Casadio, ``Tropospheric ozone
  column retrieval from esa-envisat sciamachy nadir uv/vis radiance
  measurements by means of a neural network algorithm,'' \emph{IEEE
  Transactions on Geoscience and Remote Sensing}, vol.~50, pp. 998--1011, 2012.

\bibitem{datcu2023explainable}
M.~Datcu, Z.~Huang, A.~Anghel, J.~Zhao, and R.~Cacoveanu, ``Explainable,
  physics-aware, trustworthy artificial intelligence: A paradigm shift for
  synthetic aperture radar,'' \emph{IEEE Geoscience and Remote Sensing
  Magazine}, vol.~11, no.~1, pp. 8--25, 2023.

\bibitem{kashinath2019physics}
K.~Kashinath, A.~Albert, R.~Wang, M.~Mustafa, and R.~Yu, ``Physics-informed
  spatio-temporal deep learning models,'' \emph{Bulletin of the American
  Physical Society}, vol.~64, 2019.

\bibitem{wu2018physics}
J.~Wu, K.~Kashinath, A.~Albert, M.~Prabhat, and H.~Xiao, ``Physics-informed
  generative learning to emulate unresolved physics in climate models,'' in
  \emph{American Geoscience Union Fall Meeting}, 2018.

\bibitem{camps2020advancing}
G.~Camps-Valls, M.~Reichstein, X.~Zhu, and D.~Tuia, ``{Advancing Deep Learning
  for Earth Sciences: From Hybrid Modeling to Interpretability},'' in
  \emph{International Geoscience and Remote Sensing Symposium}, 2020.

\bibitem{halevy2009unreasonable}
A.~Halevy, P.~Norvig, and F.~Pereira, ``The unreasonable effectiveness of
  data,'' \emph{IEEE Intelligent Systems}, vol.~24, no.~2, pp. 8--12, 2009.

\bibitem{moreno2018methodology}
{\'A}.~Moreno-Mart{\'\i}nez, G.~Camps-Valls, J.~Kattge, N.~Robinson,
  M.~Reichstein, P.~van Bodegom, K.~Kramer, J.~H.~C. Cornelissen, P.~Reich,
  M.~Bahn \emph{et~al.}, ``A methodology to derive global maps of leaf traits
  using remote sensing and climate data,'' \emph{Remote Sensing of
  Environment}, vol. 218, pp. 69--88, 2018.

\bibitem{Svendsen19amogape}
D.~H. Svendsen, L.~Martino, and G.~Camps-Valls, ``Active emulation of computer
  codes with gaussian processes -- application to remote sensing,''
  \emph{Pattern Recognition}, vol. 100, no. 107103, pp. 1--12, 2020.

\bibitem{ye2015equation}
H.~Ye, R.~J. Beamish, S.~M. Glaser, S.~C. Grant, C.-h. Hsieh, L.~J. Richards,
  J.~T. Schnute, and G.~Sugihara, ``Equation-free mechanistic ecosystem
  forecasting using empirical dynamic modeling,'' \emph{Proceedings of the
  National Academy of Sciences}, vol. 112, no.~13, pp. E1569--e1576, 2015.

\bibitem{daniels2015automated}
B.~C. Daniels and I.~Nemenman, ``Automated adaptive inference of
  phenomenological dynamical models,'' \emph{Nature Communications}, vol.~6, p.
  8133, 2015.

\bibitem{brunton2016discovering}
S.~L. Brunton, J.~L. Proctor, and J.~N. Kutz, ``Discovering governing equations
  from data by sparse identification of nonlinear dynamical systems,''
  \emph{Proceedings of the National Academy of Sciences}, vol. 113, no.~15, pp.
  3932--3937, 2016.

\bibitem{filchev_lachezar_2018_2475063}
L.~Filchev, L.~Pashova, V.~Kolev, and S.~Frye, ``Challenges and solutions for
  utilizing earth observations in the ``big data'' era,'' in \emph{BigSkyEarth
  Conference}, 2018.

\bibitem{tuia09active}
D.~{Tuia}, F.~Ratle, F.~Pacifici, M.~Kanevski, and W.~J. Emery, ``Active
  learning methods for remote sensing image classification,'' \emph{IEEE
  Transactions on Geoscience and Remote Sensing}, vol. 47(7), pp. 2218--2232,
  2009.

\bibitem{demir2013definition}
B.~Demir, L.~Minello, and L.~Bruzzone, ``Definition of effective training sets
  for supervised classification of remote sensing images by a novel
  cost-sensitive active learning method,'' \emph{IEEE Transactions on
  Geoscience and Remote Sensing}, vol.~52, no.~2, pp. 1272--1284, 2013.

\bibitem{tuia13learning}
D.~{Tuia} and J.~Mu{\~n}oz-Mar{\'\i}, ``Learning user's confidence for active
  learning,'' \emph{IEEE Transactions on Geoscience and Remote Sensing},
  vol.~51, no.~2, pp. 872--880, 2013.

\bibitem{rodriguez2021mapping}
A.~C. Rodr{\'\i}guez, S.~D'Aronco, K.~Schindler, and J.~D. Wegner, ``Mapping
  oil palm density at country scale: An active learning approach,''
  \emph{Remote Sensing of Environment}, vol. 261, p. 112479, 2021.

\bibitem{deepAL}
P.~Ren, Y.~Xiao, X.~Chang, P.-Y. Huang, Z.~Li, B.~B. Gupta, X.~Chen, and
  X.~Wang, ``A survey of deep active learning,'' \emph{ACM Computing Surveys},
  vol.~54, no.~9, 2021.

\bibitem{bashmal2023language}
L.~Bashmal, Y.~Bazi, F.~Melgani, M.~M. Al~Rahhal, and M.~A. Al~Zuair,
  ``Language integration in remote sensing: Tasks, datasets, and future
  directions,'' \emph{IEEE Geoscience and Remote Sensing Magazine}, vol.~11,
  no.~4, pp. 63 -- 93, 2023.

\bibitem{antol2015vqa}
S.~Antol, A.~Agrawal, J.~Lu, M.~Mitchell, D.~Batra, C.~L. Zitnick, and
  D.~Parikh, ``{VQA:} visual question answering,'' in \emph{International
  Conference on Computer Vision, ICCV}, 2015, pp. 2425--2433.

\bibitem{lobry2020rsvqa}
S.~Lobry, D.~Marcos, J.~Murray, and D.~{Tuia}, ``Rsvqa: visual question
  answering for remote sensing data,'' \emph{IEEE Transactions on Geoscience
  and Remote Sensing}, vol.~58, no.~12, pp. 8555--8566, 2020.

\bibitem{yuan2022}
Z.~Yuan, L.~Mou, Z.~Xiong, and X.~X. Zhu, ``Change detection meets visual
  question answering,'' \emph{IEEE Transactions on Geoscience and Remote
  Sensing}, vol.~60, pp. 1--13, 2022.

\bibitem{wen2023visionlanguage}
C.~Wen, Y.~Hu, X.~Li, Z.~Yuan, and X.~X. Zhu, ``Vision-language models in
  remote sensing: Current progress and future trends,'' \emph{arXiv.org
  preprint arXiv:2305.05726}, 2023.

\bibitem{Cha22pres}
C.~Chappuis, V.~Zermatten, S.~Lobry, B.~{Le Saux}, and D.~{Tuia},
  ``Prompt--rsvqa: Prompting visual context to a language model for remote
  sensing visual question answering,'' in \emph{Computer Vision and Pattern
  Recognition, CVPR Workshops}, 2022.

\bibitem{sumbul2020sd}
G.~Sumbul, S.~Nayak, and B.~Demir, ``Sd-rsic: Summarization-driven deep remote
  sensing image captioning,'' \emph{IEEE Transactions on Geoscience and Remote
  Sensing}, vol.~59, no.~8, pp. 6922--6934, 2020.

\bibitem{Kandala2022}
H.~Kandala, S.~Saha, B.~Banerjee, and X.~X. Zhu, ``Exploring transformer and
  multilabel classification for remote sensing image captioning,'' \emph{IEEE
  Geoscience and Remote Sensing Letters}, vol.~19, pp. 1--5, 2022.

\bibitem{radford2021learning}
A.~Radford, J.~W. Kim, C.~Hallacy, A.~Ramesh, G.~Goh, S.~Agarwal, G.~Sastry,
  A.~Askell, P.~Mishkin, J.~Clark \emph{et~al.}, ``Learning transferable visual
  models from natural language supervision,'' in \emph{International conference
  on machine learning}.\hskip 1em plus 0.5em minus 0.4em\relax PMLR, 2021, pp.
  8748--8763.

\bibitem{mall2023remote}
U.~Mall, C.~P. Phoo, M.~K. Liu, C.~Vondrick, B.~Hariharan, and K.~Bala,
  ``Remote sensing vision-language foundation models without annotations via
  ground remote alignment,'' \emph{arXiv preprint arXiv:2312.06960}, 2023.

\bibitem{zermatten2023text}
V.~Zermatten, J.~C. Navarro, L.~Hughes, and D.~Tuia, ``Text as a richer source
  of super- vision in semantic segmentation tasks,'' in \emph{International
  Geoscience and Remote Sensing Symposium, IGARSS}, 2023.

\bibitem{yuan2023rrsis}
Z.~Yuan, L.~Mou, Y.~Hua, and X.~X. Zhu, ``Rrsis: Referring remote sensing image
  segmentation,'' \emph{arXiv.org preprint arXiv:2306.08625}, 2023.

\bibitem{glove}
J.~Pennington, R.~Socher, and C.~Manning, ``{G}lo{V}e: Global vectors for word
  representation,'' in \emph{Proceedings of the 2014 Conference on Empirical
  Methods in Natural Language Processing, {EMNLP} 2014}.\hskip 1em plus 0.5em
  minus 0.4em\relax {ACL}, 2014, pp. 1532--1543.

\bibitem{sacha2017human}
D.~Sacha, M.~Sedlmair, L.~Zhang, J.~A. Lee, D.~Weiskopf, S.~North, and D.~Keim,
  ``Human-centered machine learning through interactive visualization,''
  \emph{Neurocomputing}, vol. 268, pp. 164--175, 2017.

\bibitem{babaee2016immersive}
M.~Babaee, S.~Tsoukalas, G.~Rigoll, and M.~Datcu, ``Immersive visualization of
  visual data using nonnegative matrix factorization,'' \emph{Neurocomputing},
  vol. 173, pp. 245--255, 2016.

\bibitem{mi2022knowledge}
L.~Mi, S.~Li, C.~Chappuis, and D.~Tuia, ``Knowledge-aware cross-modal
  text-image retrieval for remote sensing images,'' in \emph{Proceedings of the
  Second Workshop on Complex Data Challenges in Earth Observation {(CDCEO}
  2022) co-located with 31st International Joint Conference on Artificial
  Intelligence and the 25th European Conference on Artificial Intelligence
  {(IJCAI-ECAI} 2022)}, ser. {CEUR} Workshop Proceedings, vol. 3207.\hskip 1em
  plus 0.5em minus 0.4em\relax CEUR-WS.org, 2022, pp. 14--20.

\bibitem{RNY003}
A.~Jobin, M.~Ienca, and E.~Vayena, ``The global landscape of ai ethics
  guidelines,'' \emph{Nature Machine Intelligence}, vol.~1, p. 389–399, 2019.

\bibitem{kochupillai2022earth}
M.~Kochupillai, M.~Kahl, M.~Schmitt, H.~Taubenb{\"o}ck, and X.~X. Zhu, ``Earth
  observation and artificial intelligence: Understanding emerging ethical
  issues and opportunities,'' \emph{IEEE Geoscience and Remote Sensing
  Magazine}, 2022.

\bibitem{9442932}
G.~Cheng, X.~Sun, K.~Li, L.~Guo, and J.~Han, ``Perturbation-seeking generative
  adversarial networks: A defense framework for remote sensing image scene
  classification,'' \emph{IEEE Transactions on Geoscience and Remote Sensing},
  vol.~60, pp. 1--11, 2022.

\bibitem{10119208}
X.~Sun, G.~Cheng, L.~Pei, H.~Li, and J.~Han, ``Threatening patch attacks on
  object detection in optical remote sensing images,'' \emph{IEEE Transactions
  on Geoscience and Remote Sensing}, vol.~61, pp. 1--10, 2023.

\bibitem{RNY004}
%\BIBentryALTinterwordspacing
E.~Commission, ``Ethics guidlines for trustworthy ai: High-level expert group
  on artificial intelligence,'' 2018. [Online]. Available:
  \url{https://wayback.archive-it.org/12090/20201227221227/https://ec.europa.eu/digital-single-market/en/news/ethics-guidelines-trustworthy-ai}
%\BIBentrySTDinterwordspacing

\bibitem{hagendorff2020ethics}
T.~Hagendorff, ``The ethics of ai ethics: An evaluation of guidelines,''
  \emph{Minds and machines}, vol.~30, no.~1, pp. 99--120, 2020.

\bibitem{thomas2015deontology}
A.~J. Thomas, ``Deontology, consequentialism and moral realism.''
  \emph{Minerva: An Internet Journal of Philosophy}, vol.~19, 2015.

\bibitem{RNY005}
M.~Kochupillai, ``Outline of a novel approach for identifying ethical issues in
  early stages of ai4eo research,'' \emph{2021 IEEE International Geoscience
  and Remote Sensing Symposium IGARSS}, pp. 165--1168, 2021.

\bibitem{llmethics}
K.~Bugbee and R.~Ramachandran, ``The ethics of large language models: Who
  controls the future of open science?''
  https://www.earthdata.nasa.gov/learn/blog/ethics-large-language-models, 2023.

\bibitem{dataeth2018de}
B.~d.~I. Daten Ethik~Kommission, ``Opinion of the data ethics commission -
  executive summary,''
  https://www.bfdi.bund.de/SharedDocs/Downloads/EN/Datenschutz/Data-Ethics-Commission\_Opinion.pdf?\_\_blob=publicationFile\&v=1,
  p.~32, 2018.

\bibitem{RNY008}
COPA, COGECA, CEMA, F.~Europe, CEETAR, ECPA, EFFAB, FEFAC, and ESA, ``Eu code
  of conduct on agricultural data sharing by contractual agreement,''
  \url{https://www.copa-cogeca.eu/img/user/files/EU\%20CODE/EU\_Code\_2018\_web\_version.pdf},
  2018.

\bibitem{rs12223803}
R.~Schneider, A.~M. Vicedo-Cabrera, F.~Sera, P.~Masselot, M.~Stafoggia,
  K.~de~Hoogh, I.~Kloog, S.~Reis, M.~Vieno, and A.~Gasparrini, ``A
  satellite-based spatio-temporal machine learning model to reconstruct daily
  pm2.5 concentrations across great britain,'' \emph{Remote Sensing}, vol.~12,
  no.~22, 2020.

\bibitem{atmos11030239}
M.~Stafoggia, C.~Johansson, P.~Glantz, M.~Renzi, A.~Shtein, K.~de~Hoogh,
  I.~Kloog, M.~Davoli, P.~Michelozzi, and T.~Bellander, ``A random forest
  approach to estimate daily particulate matter, nitrogen dioxide, and ozone at
  fine spatial resolution in sweden,'' \emph{Atmosphere}, vol.~11, no.~3, 2020.

\bibitem{CHEN2019104934}
J.~Chen, K.~{de Hoogh}, J.~Gulliver, B.~Hoffmann, O.~Hertel, M.~Ketzel,
  M.~Bauwelinck, A.~{van Donkelaar}, U.~A. Hvidtfeldt, K.~Katsouyanni, N.~A.
  Janssen, R.~V. Martin, E.~Samoli, P.~E. Schwartz, M.~Stafoggia, T.~Bellander,
  M.~Strak, K.~Wolf, D.~Vienneau, R.~Vermeulen, B.~Brunekreef, and G.~Hoek, ``A
  comparison of linear regression, regularization, and machine learning
  algorithms to develop europe-wide spatial models of fine particles and
  nitrogen dioxide,'' \emph{Environment International}, vol. 130, p. 104934,
  2019.

\bibitem{DI2019104909}
Q.~Di, H.~Amini, L.~Shi, I.~Kloog, R.~Silvern, J.~Kelly, M.~B. Sabath,
  C.~Choirat, P.~Koutrakis, A.~Lyapustin, Y.~Wang, L.~J. Mickley, and
  J.~Schwartz, ``An ensemble-based model of pm2.5 concentration across the
  contiguous united states with high spatiotemporal resolution,''
  \emph{Environment International}, vol. 130, p. 104909, 2019.

\bibitem{schneider2022}
R.~Schneider, P.~Masselot, A.~M. Vicedo-Cabrera, F.~Sera, M.~Blangiardo,
  C.~Forlani, J.~Douros, O.~Jorba, M.~Adani, R.~Kouznetsov, F.~Couvidat,
  J.~Arteta, B.~Raux, M.~Guevara, A.~Colette, J.~Barr\'{e}, V.-H. Peuch, and
  A.~Gasparrini, ``Differential impact of government lockdown policies on
  reducing air pollution levels and related mortality in europe,''
  \emph{Scientific Reports}, vol.~12, no.~1, p. 726, 2022.

\bibitem{vicedo2021burden}
A.~M. Vicedo-Cabrera, N.~Scovronick, F.~Sera, D.~Roy{\'e}, R.~Schneider,
  A.~Tobias, C.~Astrom, Y.~Guo, Y.~Honda, D.~Hondula \emph{et~al.}, ``The
  burden of heat-related mortality attributable to recent human-induced climate
  change,'' \emph{Nature climate change}, vol.~11, no.~6, pp. 492--500, 2021.

\bibitem{URBAN2021111227}
A.~Urban, C.~{Di Napoli}, H.~L. Cloke, J.~Kysel\'{y}, F.~Pappenberger, F.~Sera,
  R.~Schneider, A.~M. Vicedo-Cabrera, F.~Acquaotta, M.~S. Ragettli, C.~I.
  {n}iguez, A.~Tobias, E.~Indermitte, H.~Orru, J.~J. Jaakkola, N.~R. Ryti,
  M.~Pascal, V.~Huber, A.~Schneider, F.~{de' Donato}, P.~Michelozzi, and
  A.~Gasparrini, ``Evaluation of the era5 reanalysis-based universal thermal
  climate index on mortality data in europe,'' \emph{Environmental Research},
  vol. 198, p. 111227, 2021.

\end{thebibliography}
